\documentclass[conference]{IEEEtran}
\usepackage{times}

\usepackage[numbers]{natbib}
\usepackage{multicol}
\usepackage{xcolor}
\usepackage[bookmarks=true]{hyperref}

\usepackage{booktabs}
\usepackage{algorithm}
\usepackage{algorithmic}
\usepackage{amsmath}
\usepackage{amssymb}
\usepackage{bm}  
\usepackage{graphicx}
\usepackage{colortbl}  
\usepackage{multirow}  
\usepackage{stfloats}  
\usepackage{caption}  
\usepackage{cuted}  
\usepackage{listings}  

\lstdefinestyle{promptboxstyle}{
  basicstyle=\ttfamily\footnotesize,
  breaklines=true,
  breakatwhitespace=false,
  frame=single,
  framerule=0.35pt,
  rulecolor=\color{black!20},
  backgroundcolor=\color{black!1},
  framesep=3pt,
  aboveskip=0.6em,
  belowskip=0.6em,
  xleftmargin=0pt,
  xrightmargin=0pt,
  showstringspaces=false,
  columns=fullflexible,
  keepspaces=true,
  tabsize=2,
  extendedchars=true,
  inputencoding=utf8,
  literate={->}{{$\rightarrow$}}2
           {á}{{\'a}}1 {é}{{\'e}}1 {í}{{\'i}}1 {ó}{{\'o}}1 {ú}{{\'u}}1
}
\lstnewenvironment{promptbox}{\lstset{style=promptboxstyle,linewidth=\columnwidth}}{}

\newenvironment{vlmbox}{%
  \par\vspace{0.5em}%
  \begin{quote}%
  \small\itshape%
}{%
  \end{quote}%
  \vspace{0.5em}%
}

\newcommand{\method}{\textsc{Vid2Sid}}
\newcommand{\simtoreal}{\texttt{sim2real}}
\newcommand{\simtosim}{\texttt{sim2sim}}

\newcommand{\apptocmain}[2]{%
  \vspace{8pt}%
  \noindent{\large\bfseries #1}\dotfill{\large\bfseries\pageref{#2}}\par\vspace{1pt}}
\newcommand{\apptocsub}[2]{%
  \noindent\hspace{2em}#1\dotfill\pageref{#2}\par\vspace{1pt}}
\newif\ifshowcomments
\showcommentsfalse

\let\labelindent\relax
\usepackage{enumitem}
\setlist{nosep, leftmargin=*}
\setlist[enumerate]{label=\arabic*)}

\begin{document}

\title{\method{}: Videos Can Help Close the Sim2Real Gap}

\author{%
\authorblockN{Kevin Qiu$^{1,2}$,
Yu Zhang$^{3}$,
Marek Cygan$^{1,4}$,
Josie Hughes$^{3}$}
\authorblockA{$^{1}$University of Warsaw \quad
$^{2}$IDEAS NCBR \quad
$^{3}$EPFL \quad
$^{4}$Nomagic}
\authorblockA{\texttt{kevinxqiu@gmail.com}}}

\maketitle

\begin{strip}
\vspace{-1cm}
\begin{center}
\includegraphics[width=0.98\textwidth]{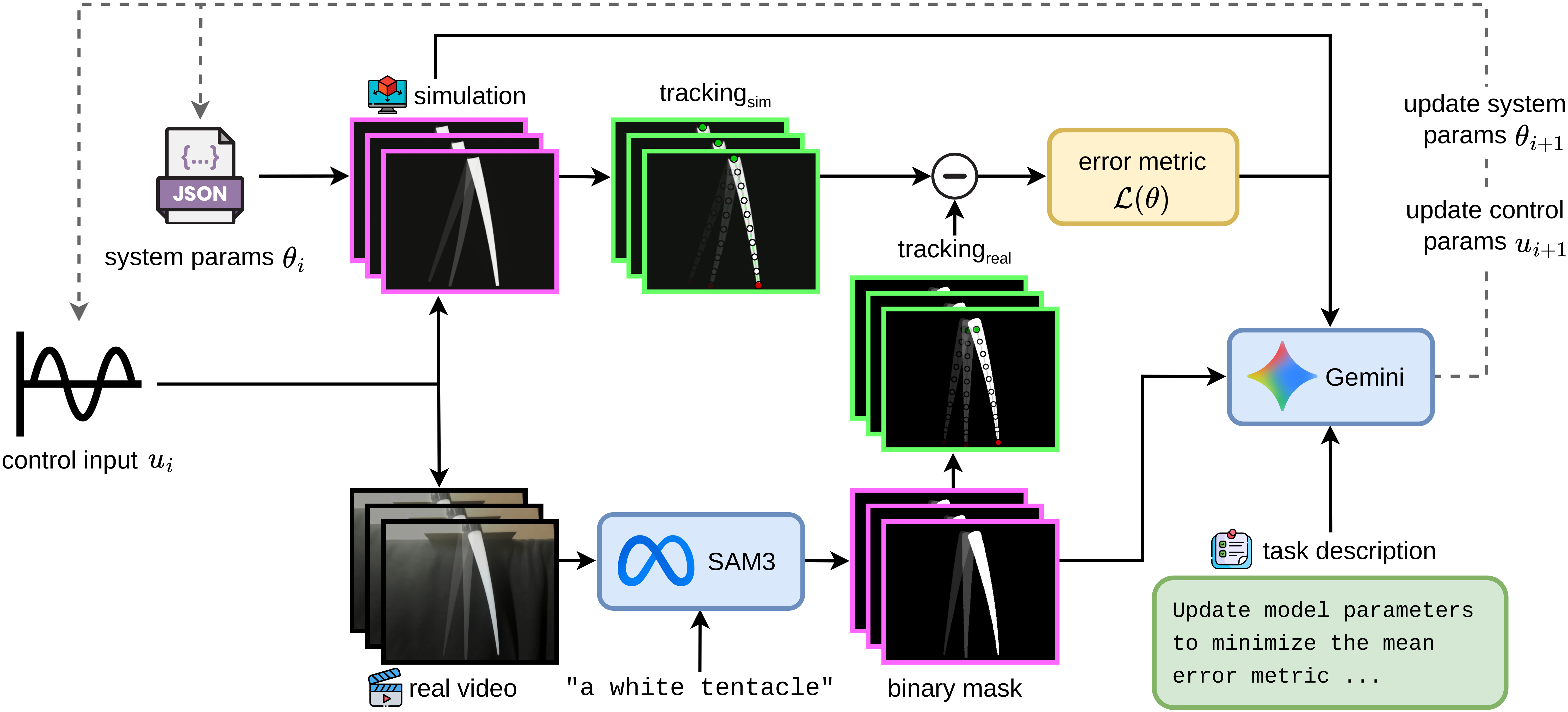}
\captionof{figure}{Overview of \method{}. Given paired sim-real videos, the \emph{perception layer} extracts trajectories (SAM3 centerlines for soft robots, marker tracking for rigid robots), and the \emph{reasoning layer} uses a VLM to diagnose discrepancies and propose physics parameter updates with natural language rationales. This closed-loop process matches or exceeds black-box optimizers within 10 iterations while requiring no task-specific training or optimizer hyperparameters.}
\label{fig:framework}
\end{center}
\end{strip}

\begin{abstract}
Calibrating a robot simulator's physics parameters (friction, damping, material stiffness) to match real hardware is often done by hand or with black-box optimizers that reduce error but cannot explain \emph{which} physical discrepancies drive the error. When sensing is limited to external cameras, the problem is further compounded by perception noise and the absence of direct force or state measurements.
We present \method{}, a video-driven system identification pipeline that couples foundation-model perception with a VLM-in-the-loop optimizer that analyzes paired sim-real videos, diagnoses concrete mismatches, and proposes physics parameter updates with natural language rationales.
We evaluate our approach on a tendon-actuated finger (rigid-body dynamics in MuJoCo) and a deformable continuum tentacle (soft-body dynamics in PyElastica).
On \simtoreal{} holdout controls unseen during training, \method{} achieves the best average rank across all settings, matching or exceeding black-box optimizers while uniquely providing interpretable reasoning at each iteration.
\simtosim{} validation confirms that \method{} recovers ground-truth parameters most accurately (mean relative error under 13\% vs.\ 28--98\%), and ablation analysis reveals three calibration regimes.
VLM-guided optimization excels when perception is clean and the simulator is expressive, while model-class limitations bound performance in more challenging settings.
\end{abstract}

\IEEEpeerreviewmaketitle
\vspace{-0.5cm}
\section{Introduction}
\label{sec:introduction}

Simulation has become the dominant paradigm for training robot control policies, enabling thousands of trials without wear on physical hardware~\citep{andrychowicz2020learning}.
Yet deploying simulation-trained policies on real robots, the \simtoreal{} transfer problem, remains a fundamental challenge~\citep{muratore2022robot}.
When the simulator's physics parameters (joint friction, material stiffness, damping coefficients) do not match the physical hardware, policies that perform flawlessly in simulation can fail on contact with the real world.
Left unresolved, this physics gap blocks the transfer of any simulation-trained policy to the real world, from rigid manipulators to soft continuum robots.

Two broad strategies have emerged to close this gap.
Domain randomization~\citep{tobin2017domain, peng2018sim, andrychowicz2020learning} trains policies robust to parameter variation, sidestepping the need to identify specific physics parameters.
While effective in some settings, it sacrifices optimality on any particular configuration, requires conservatively wide parameter ranges that can degrade policy quality, and provides no insight into the actual physical properties of the hardware.
The alternative, directly identifying the simulator's physics parameters to build a faithful digital twin, is more fundamental. An accurately identified simulator enables not only robust policy transfer but also downstream debugging, iterative design, and reuse across tasks.
But identification is an inherently difficult inverse problem. The mapping from physics parameters to observed motion is nonlinear, different parameter combinations can produce indistinguishable behaviors, and real-world observations are corrupted by perception noise and unmodeled effects.
Existing methods offer unsatisfying options: manual tuning by domain experts, which is slow and unscalable, or black-box optimization (Bayesian optimization~\citep{calandra2016bayesian}, CMA-ES, SimOpt~\citep{chebotar2019closing}), which automates parameter search but provides no insight into \emph{which} physical discrepancies drive error~\citep{allevato2020tunenet}.

Meanwhile, foundation models have transformed robot perception (SAM3~\citep{carion2025sam3segmentconcepts} enables markerless video tracking without task-specific training), and VLMs can reason about physical dynamics from video~\citep{wiedemer2025video}.
Yet no prior work has applied these capabilities to quantitative physics parameter identification, the very task where visual reasoning about dynamics should be most informative. VLM-based robotics has focused on task planning~\citep{ahn2022saycan}, reward design~\citep{ma2023eureka}, and scene-level domain randomization~\citep{yu2024lang4sim2real}, leaving low-level system identification untouched.

We present \method{}, a \underline{vid}eo-\underline{to}-\underline{s}ystem-\underline{id}entification pipeline that bridges this gap by coupling foundation-model perception with VLM-guided optimization (Figure~\ref{fig:framework}).
\method{} decomposes into two layers. The \emph{perception layer} extracts trajectories from video (SAM3 centerlines for soft robots, marker tracking for rigid robots). The \emph{reasoning layer} uses a VLM to analyze paired sim-real videos, diagnose specific discrepancies (e.g., ``the simulated tentacle oscillates too fast''), and propose physics parameter updates with natural language rationales.
Unlike black-box methods, \method{} provides interpretable diagnostics at every iteration and requires no optimizer hyperparameters such as acquisition functions, step sizes, or population sizes.

Our key contributions are as follows:
\begin{enumerate}
    \item \textbf{Video as a modality for \simtoreal{} reasoning.} We validate that VLMs can reason about physics discrepancies from paired sim-real video across morphologically distinct embodiments, diagnosing specific causes of the \simtoreal{} gap (e.g., overdamped joints, incorrect material stiffness) with natural language rationales.

    \item \textbf{Video-driven system identification pipeline.} We introduce \method{}, a two-layer system combining foundation model perception with VLM-guided parameter optimization that works across different physical systems and simulation backends without requiring differentiable simulators, task-specific training, or manual annotation.

    \item \textbf{Cross-morphology demonstration on unseen control profiles.} We evaluate \method{} on a tendon-actuated finger (MuJoCo) and a continuum tentacle (PyElastica), testing generalization to holdout control profiles unseen during training. \method{} achieves the best average rank across all \simtoreal{} settings and recovers ground-truth parameters most accurately in \simtosim{} (under 13\% relative error vs.\ 28--98\% for baselines).
\end{enumerate}

\section{Related Work}
\label{sec:related_work}

\subsection{\simtoreal{} Transfer and System Identification}

The \simtoreal{} gap has motivated extensive work on domain adaptation and system identification~\citep{muratore2022robot}.
Domain randomization~\citep{tobin2017domain, peng2018sim}, scaled to dexterous manipulation by \citet{andrychowicz2020learning}, trains policies robust to parameter variation rather than identifying parameters directly.
Iterative calibration methods adapt parameters from real-world experience. SimOpt~\citep{chebotar2019closing} adjusts simulation distributions, \citet{du2021auto} predict parameter updates from pixels, BayesSim~\citep{ramos2019bayessim} performs probabilistic inference, and TuneNet~\citep{allevato2020tunenet} learns residual corrections.
Bayesian optimization~\citep{calandra2016bayesian} and differentiable simulation~\citep{jatavallabhula2021gradsim, dubied2022sim} offer sample efficiency and gradient-based identification, respectively.
Differentiable approaches are powerful when an analytical gradient through the simulator exists, but many simulators (including PyElastica's Cosserat rod solver) are non-differentiable or require custom adjoint implementations.
VLM-guided optimization sidesteps this requirement entirely by operating on rendered video, making it simulator-agnostic and applicable to any physics engine that produces visual output.
For soft robots, material properties are difficult to measure \emph{in situ}~\citep{george2018control}. Our work addresses both rigid and soft regimes using visual feedback without requiring explicit measurements or differentiable physics.

\subsection{Foundation Models for Robot Perception}

Robot perception traditionally relies on physical markers or task-specific learned models~\citep{wang2019visual, mathis2018deeplabcut}, which constrain the workspace or require training data.
Foundation models have shifted this landscape. SAM~\citep{kirillov2023sam} demonstrated zero-shot segmentation, and SAM3~\citep{carion2025sam3segmentconcepts} extends this to video via text or point prompts, which is critical for soft robots where appearance varies widely.
We leverage SAM3 for markerless tentacle tracking and a simple marker-based tracker for the finger, enabling rapid deployment on morphologically diverse platforms.

\subsection{VLMs for Physical Reasoning in Robotics}

Large language models have been applied to robotics for task planning~\citep{ahn2022saycan}. Vision-language-action models enable direct visuomotor control~\citep{zitkovich2023rt, driess2023palm}, and LLM-driven reward design~\citep{ma2023eureka} automates reward specification for reinforcement learning.
LLMs have also been applied to robot co-design, evolving morphologies~\citep{qiu2024robomorph} and mediating design trade-offs through structured debate~\citep{qiu2025debate2create}.
\citet{yu2024lang4sim2real} demonstrated that natural language descriptions can guide domain randomization for manipulation.
\citet{patel2025real} proposed using VLMs to generate keypoint-based rewards, iteratively refining reward functions from visual observations.

Most closely related, \citet{wiedemer2025video} showed that video models can reason zero-shot about physical dynamics.
\method{} operationalizes this insight for low-level system identification. Rather than planning or reward specification, it directly maps visual observations to numerical parameter recommendations while respecting parameter bounds and optimization history.

\section{\method{}}
\label{sec:method}

\method{} identifies a simulator's physics parameters through an iterative closed-loop process.
At each iteration, the same control signal is sent to both the simulator and the real robot, and video of both executions is recorded.
A perception system extracts structured trajectories from the videos, and a VLM analyzes the paired recordings to diagnose physical discrepancies and propose parameter updates.
This loop repeats until simulated behavior converges to real behavior.

Concretely, the pipeline decomposes into two layers.
The \textbf{perception layer} extracts trajectory representations from raw video using a task-appropriate perception function.
The \textbf{reasoning layer} uses a VLM to diagnose sim-real discrepancies and propose parameter updates with natural language rationales.

\textbf{Problem formulation.}
Let $f_\theta$ denote a simulator parameterized by physics parameters $\theta \in \Theta$, where $\Theta$ defines valid bounds.
Given a control signal $u$ applied to both the simulator and the real robot, let $\tau_\text{sim}(\theta) = g(v_\text{sim}(\theta))$ and $\tau_\text{real} = g(v_\text{real})$ be trajectories extracted from the simulated and real videos, respectively, where $g$ is a perception function (SAM3 centerline extraction or marker tracking).
The objective is to find the optimal physics parameters $\theta^*$ that minimize the alignment error between sim and real:
\begin{equation}
    \theta^* = \arg\min_{\theta \in \Theta}\; \mathcal{L}\!\left(\tau_\text{sim}(\theta),\; \tau_\text{real}\right)
\end{equation}
where $\mathcal{L}$ is the mean absolute error between temporally aligned trajectories (Section~\ref{sec:metrics}).
\method{} solves this problem by using a VLM to propose successive parameter updates based on paired sim-real videos, error metrics, and optimization history.

\subsection{Simulation Framework}
\label{sec:simulation}

\begin{figure}[!t]
    \centering
    \includegraphics[width=\columnwidth]{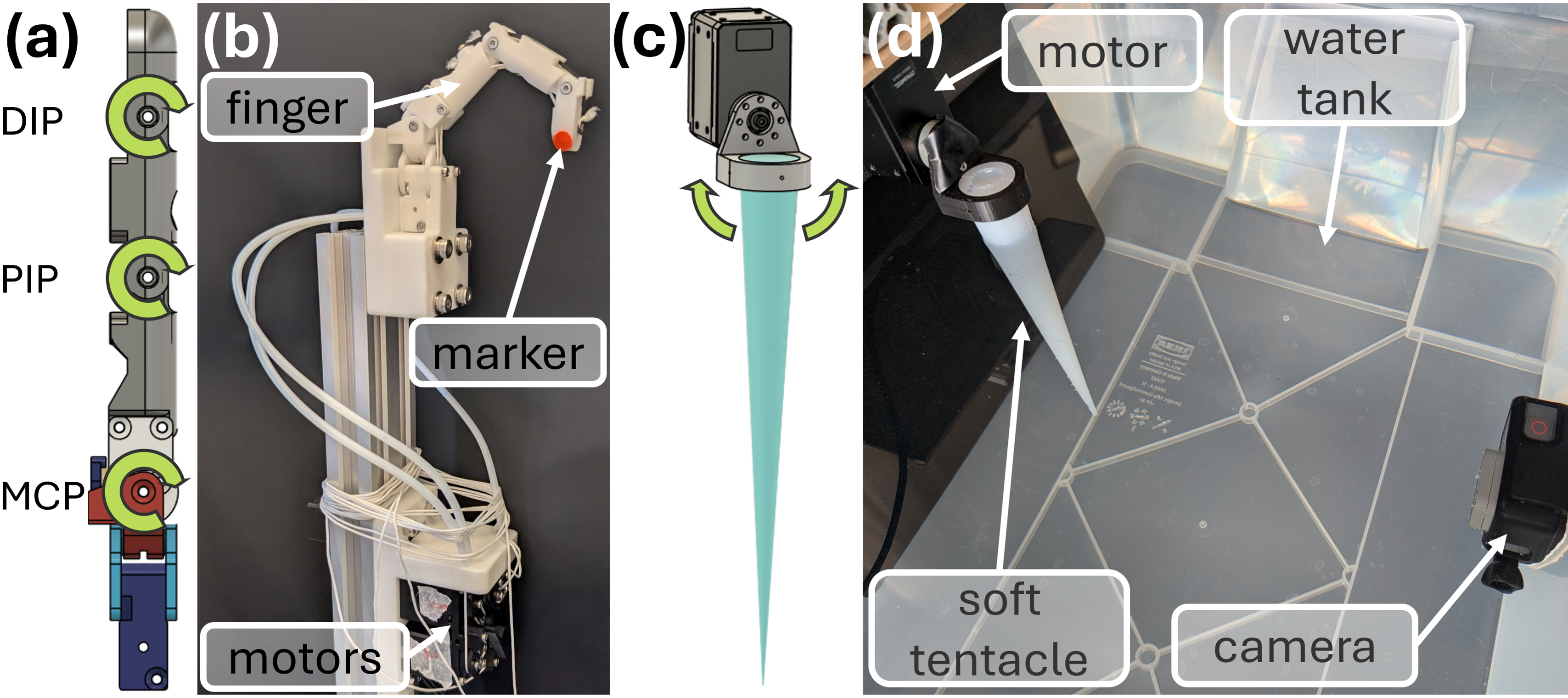}
    \caption{Experimental platforms. (a)~CAD model of tendon-driven finger. (b)~Physical finger with tracked marker. (c)~CAD model of soft tentacle. (d)~Underwater tentacle setup. These platforms span rigid-body and continuum dynamics, enabling evaluation across qualitatively different calibration regimes.}
    \label{fig:setup}
\end{figure}

\method{} is simulator-agnostic, requiring only that the simulator produce visual output.
We evaluate on two physical platforms, a tendon-actuated finger and a soft continuum tentacle (Figure~\ref{fig:setup}), using two simulation backends representing qualitatively different dynamics regimes.

\subsubsection{MuJoCo (Rigid-Body Dynamics)}

The finger is a three-link kinematic chain modeled in MuJoCo. Joints use PD position control with fixed gain ($k_p{=}1000$~N/rad) and tunable damping $k_d$.

\textbf{Tunable parameters:} joint friction, viscous damping, rotor inertia, and link density (parameter bounds are provided in Appendix~\ref{app:bounds}).

\subsubsection{PyElastica (Continuum Dynamics)}

The tentacle is modeled as a Cosserat rod~\citep{gazzola2018forward} using PyElastica~\citep{PyElastica}. This framework is well-suited for continuum and soft robots.
The rod is subject to elastic restoring forces parameterized by Young's modulus $E$ and Poisson ratio $\nu$, a fixed-base boundary condition with prescribed oscillatory rotation, linear velocity damping with coefficient $\gamma$, and external forces including gravity and (for underwater settings) hydrodynamic interaction forces.

\textbf{Tunable body parameters} (used for in-air calibration): Young's modulus $E$, rod density $\rho$, Poisson ratio $\nu$, and damping coefficient $\gamma$. These are four material parameters with bounds in Appendix~\ref{app:bounds}.

\textbf{Tunable environment parameters} (used for underwater calibration):
fluid density $\rho_f$, perpendicular drag coefficient $C_\perp$, and tangential drag coefficient $C_\parallel$, modeling anisotropic hydrodynamic drag~\citep{gazzola2018forward} (bounds in Appendix~\ref{app:bounds}).
In the \textbf{in-air} setting, we tune body parameters only. In the \textbf{underwater stress test}, we freeze body parameters and tune only environment parameters (Section~\ref{sec:exp_water}).
In both settings, the VLM can also adjust control parameters (motor amplitudes) to elicit more informative motions (Section~\ref{sec:active_learning}).

\subsection{Control Signal Generation}
\label{sec:control}

Both platforms use sinusoidal actuation $u(t) = A \sin(2\pi f t + \phi)$. Sinusoids provide repeatable, bounded motion that exercises the full dynamic range of each joint. Differences in oscillation amplitude, phase, and settling time reveal the underlying physical parameters, making sinusoids an informative and easily parameterized test signal.
For the tentacle, a single motor drives base oscillation at fixed frequency (0.15\,Hz during training; holdout profiles vary frequency and amplitude).
For the finger, two joints are actuated with independent sinusoids.
The same control signal is sent to both simulator and real hardware. Full control parameterization details are provided in Appendix~\ref{app:bounds}.

\subsection{Real-World Capture}
\label{sec:capture}

The real robot is actuated by Dynamixel servos receiving position commands at a fixed rate. A webcam captures video at 30 fps with controlled exposure.
Each trial follows a standardized timing protocol: initialize to home position, hold for 2 seconds, execute motion for 10 seconds while recording video, then save with timestamps.
Hardware specifications (motor models, camera settings, compute resources) are detailed in Appendix~\ref{app:hardware}.

\subsection{Trajectory Extraction from Video}
\label{sec:segmentation}

\subsubsection{Tentacle (SAM3 Segmentation + Centerline Extraction)}

\begin{figure}[!t]
    \centering
    \includegraphics[width=\columnwidth]{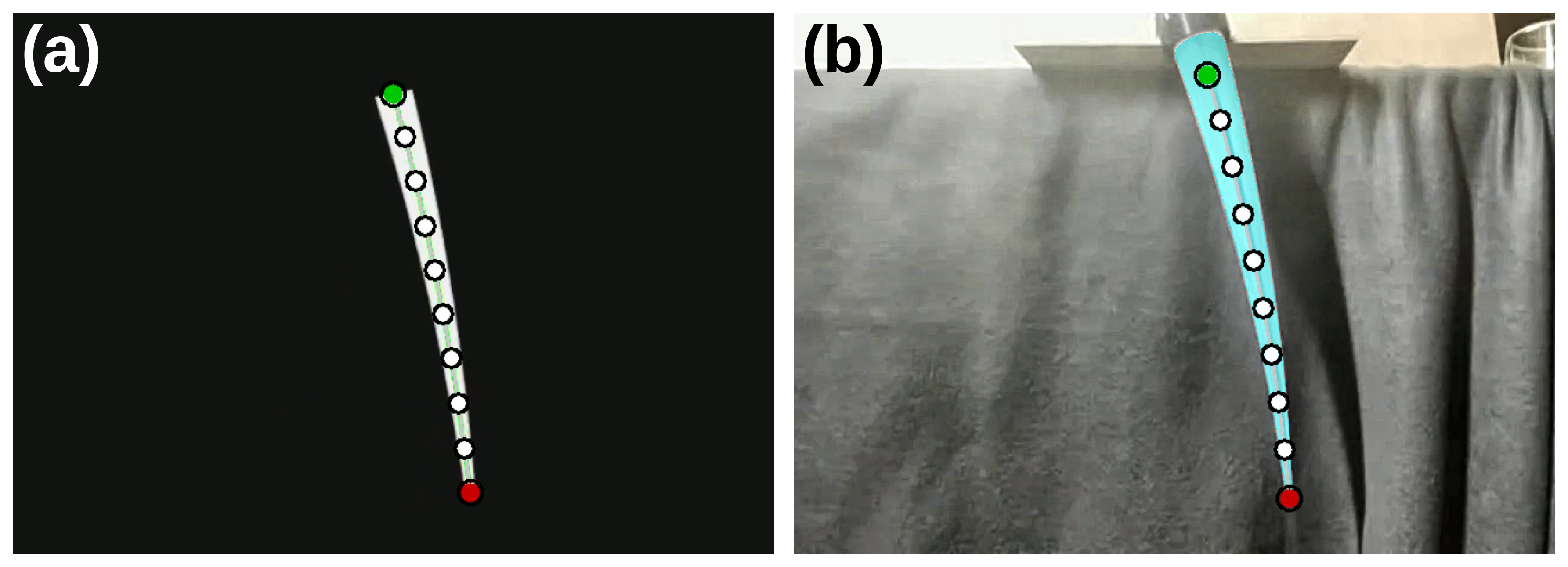}
    \caption{Centerline extraction pipeline. (a)~Simulated centerline (10 equidistant points, base to tip). (b)~Real video frame with SAM3 segmentation mask and extracted centerline overlay. The shared 10-point representation enables direct point-wise error computation between sim and real without manual annotation.}
    \label{fig:centerline}
\end{figure}

We use SAM3~\citep{carion2025sam3segmentconcepts} for real tentacle video segmentation, prompting with a text description on the first frame and propagating through the video.
Text prompts alone achieve 97.2\% segmentation success across all videos used in our experiments, requiring no manual point prompts (Appendix~\ref{app:sam3}).
Empty masks (from blur or occlusion) are interpolated from neighboring frames.
From each binary mask, we extract a 10-point centerline by computing the medial axis and resampling along arc length (Figure~\ref{fig:centerline}), producing per-frame centerlines $\{(x_i^t, y_i^t)\}_{i=1}^{10}$. Ten points provide sufficient spatial resolution to capture the tentacle's deformation modes while remaining robust to segmentation noise. The simulator directly provides rendered centerlines for comparison. Implementation details are in Appendix~\ref{app:sam3}.

\subsubsection{Finger (Tip Extraction)}

For the finger, we track only the 2D fingertip marker position: (1)~crop to a fixed region-of-interest around the workspace, (2)~threshold in HSV color space to segment the marker, and (3)~select the largest connected component and record its centroid $(x_\text{tip}^t, y_\text{tip}^t)$ per frame.

This single-point representation is sufficient because the finger's kinematic structure constrains the configuration space.

\subsection{Error Metrics}
\label{sec:metrics}

Before computing error, sim and real trajectories are temporally aligned via cross-correlation (bounded lag window, max 1\,s) and compared in 2D pixel coordinates from matched camera viewpoints, with arc-length normalization for the tentacle.

We use mean absolute error (MAE) as the primary metric. For the tentacle, we compute point-wise MAE across all $N{=}10$ centerline points and $T$ frames:
\begin{equation}
    \text{MAE}_\text{centerline} = \frac{1}{TN} \sum_{t=1}^T \sum_{i=1}^N \left\| \mathbf{p}_i^{t,\text{sim}} - \mathbf{p}_i^{t,\text{real}} \right\|
\end{equation}
For the finger, we compute MAE over the tip trajectory:
\begin{equation}
    \text{MAE}_\text{tip} = \frac{1}{T} \sum_{t=1}^T \left\| \mathbf{p}_\text{tip}^{t,\text{sim}} - \mathbf{p}_\text{tip}^{t,\text{real}} \right\|
\end{equation}
MAE is less sensitive to outliers than RMSE. We ignore the initial 5\,s of each recording to exclude transients.

\subsection{VLM-Driven Parameter Tuning}
\label{sec:vlm}

\subsubsection{Prompt Design}

The VLM receives a multimodal prompt containing sim and real videos, current parameters $\theta$ with bounds, error metrics, and iteration history for in-context learning~\citep{brown2020language}.
An optional chain-of-thought instruction~\citep{wei2022chain} requests explicit reasoning before the final answer.
The VLM returns recommended parameter values, a self-assessed confidence score (0--1), and a natural language rationale linking observed discrepancies to specific parameter changes.
Proposed values are clamped to valid bounds.
We use Google Gemini 2.5 Pro~\citep{comanici2025gemini} with default decoding settings (temperature 1.0, top-$P$ 0.95), held fixed across all seeds and domains (Appendix~\ref{app:vlm_settings}).
Full prompt templates are provided in Appendix~\ref{app:prompts}.

\subsubsection{Iterative Loop}

Algorithm~\ref{alg:vid2sid} summarizes the optimization loop. We run for a fixed budget of $K{=}10$ iterations. To handle VLM stochasticity, we track the best parameters seen and return them at termination, even if later iterations degrade performance.

\begin{algorithm}[!t]
\caption{\method{} iterative calibration loop.}
\label{alg:vid2sid}
\begin{algorithmic}[1]
\REQUIRE Initial parameters $\theta_0$, control $u_0$, bounds $\Theta$, $\mathcal{U}$, max iterations $K$
\STATE $\theta \gets \theta_0$, \quad $u \gets u_0$, \quad $\theta^* \gets \theta_0$, \quad $\text{best\_error} \gets \infty$
\STATE $H \gets \emptyset$ \COMMENT{Iteration history}
\FOR{$k = 1, \ldots, K$}
    \STATE $v_\text{sim} \gets \text{RunSimulation}(\theta, u)$
    \STATE $v_\text{real} \gets \text{CaptureReal}(u)$
    \STATE $\tau_\text{sim}, \tau_\text{real} \gets \text{ExtractTraj}(v_\text{sim}, v_\text{real})$
    \STATE $\tau_\text{sim}, \tau_\text{real} \gets \text{Align}(\tau_\text{sim}, \tau_\text{real})$
    \STATE $\text{error} \gets \text{ComputeMAE}(\tau_\text{sim}, \tau_\text{real})$
    \IF{$\text{error} < \text{best\_error}$}
        \STATE $\text{best\_error} \gets \text{error}$, \quad $\theta^* \gets \theta$
    \ENDIF
    \STATE $H \gets H \cup \{(\theta, u, \text{error})\}$
    \STATE $\theta', u' \gets \text{Recommend}(v_\text{sim}, v_\text{real}, \theta, u, \text{error}, H)$
    \STATE $\theta \gets \text{Clamp}(\theta', \Theta)$, \quad $u \gets \text{Clamp}(u', \mathcal{U})$
\ENDFOR
\RETURN $\theta^*$
\end{algorithmic}
\end{algorithm}

\subsubsection{Active Learning over Control Inputs}
\label{sec:active_learning}

In addition to physics parameters, the VLM can optionally recommend changes to \emph{control parameters} (motor amplitudes) to elicit more informative motions. For the tentacle, amplitude ranges from 0.2--1.0\,rad with frequency fixed at 0.15\,Hz during training. For the finger, both MCP and PIP amplitudes are tunable (0--60$^\circ$).

To prevent degenerate solutions where the optimizer minimizes motion to trivially reduce pixel error, we enforce minimum amplitude bounds and evaluate on holdout controls with \emph{fixed} profiles (H1--H4). We also compare active learning against fixed-control baselines to isolate its effect (Section~\ref{sec:ablations}).

\section{Experiments}
\label{sec:experiments}

We evaluate \method{} on two platforms, an articulated finger and a continuum tentacle, across three settings: \simtosim{} validation, \simtoreal{} calibration in air, and an underwater tentacle stress test where only environment parameters are tuned (shared-body protocol).
Across all methods we enforce a fixed training budget of 10 iterations per seed.

\subsection{Experimental Setup}

\subsubsection{Hardware Configuration}

Both platforms are introduced in Section~\ref{sec:simulation} (Figure~\ref{fig:setup}).

\emph{Tendon-driven finger.}
A three-link articulated finger 3D-printed from PLA, actuated by two Dynamixel XC330-T288-T motors via Bowden cables.
The metacarpophalangeal (MCP) joint is antagonistically driven by one motor, while the proximal and distal interphalangeal (PIP/DIP) joints are coupled and driven by the second motor, with elastic return.

\emph{Soft tentacle.}
A silicone tentacle mold-cast from DragonSkin 10 Slow elastomer, attached to a Dynamixel XW430-T200R motor.
For the in-air \simtoreal{} task, we record with a static external camera against a dark background.
For the underwater stress test, the same tentacle is submerged and recorded with a high-frame-rate action camera. Details are provided in Appendix~\ref{app:hardware}.

\subsubsection{Evaluation Protocol}
\label{sec:eval_protocol}

All methods receive 10 iterations per seed (CMA-ES evaluates a full population per iteration, giving it more total function evaluations). For each seed, we select the iteration with lowest training error and evaluate its holdout MAE on 4 unseen control profiles (H1--H4) spanning corners of the control space, each repeated 3 times on hardware. For the tentacle, holdouts vary both amplitude and frequency, while training holds frequency fixed. This tests whether identified parameters transfer to unseen actuation regimes, not just unseen amplitudes. We report mean $\pm$ std over $N{=}3$ training seeds, where each seed aggregates 12 datapoints ($4$ holdouts $\times$ $3$ repeats), so reported uncertainty reflects algorithmic sensitivity rather than per-trial noise.

\subsection{\simtosim{} Calibration}
\label{sec:exp_sim2sim}

Before deploying on real hardware, we validate in a controlled \simtosim{} setting where both ``real'' and ``sim'' are simulations with different parameter values. A ``ground truth'' simulation is created with fixed parameters. Candidate simulations start from random initializations and are optimized under the same budget. \simtosim{} provides an upper bound on achievable performance by eliminating camera noise, segmentation errors, and unmodeled hardware effects.

\subsection{\simtoreal{} Calibration}
\label{sec:exp_sim2real}

For both platforms, initial parameters are randomly sampled from bounds for each training seed. All methods receive identical initializations for fair comparison. We report mean absolute 2D tip error (pixels) for the finger and mean absolute centerline error (pixels) for the tentacle, using the perception and metrics described in Section~\ref{sec:method}.

\subsection{Stress Test: Underwater Tentacle}
\label{sec:exp_water}

To probe pipeline limits when model misspecification dominates, we evaluate an underwater setting using a shared-body protocol. All methods share fixed body parameters discovered by \method{} during in-air calibration (best seed) and tune only three hydrodynamic environment parameters (fluid density, perpendicular drag coefficient, tangential drag coefficient).
Note that this protocol advantages baselines, which start from body parameters they did not discover.

This setup isolates model class from optimizer quality. Any residual sim-real gap is attributable to unmodeled hydrodynamic effects (turbulence, vortex shedding, buoyancy) rather than optimizer choice. We frame these results as a limitation study revealing when optimizer choice matters less than model expressiveness.

\subsection{Baseline Comparisons}
\label{sec:baselines}

We compare \method{} against five black-box optimization baselines, all given the same 10-iteration budget and parameter bounds:
\begin{itemize}
    \item \textbf{Random:} Uniform random sampling within bounds (lower bound baseline).
    \item \textbf{Nelder-Mead:} Derivative-free simplex method~\citep{singer2009nelder} (SciPy~\citep{virtanen2020scipy}).
    \item \textbf{Golden-CD:} Sequential 1D golden-section search along each parameter axis.
    \item \textbf{Bayesian Optimization (BO):} Gaussian Process surrogate with Expected Improvement acquisition~\citep{jones1998efficient} (scikit-optimize, default hyperparameters).
    \item \textbf{CMA-ES:} Covariance Matrix Adaptation Evolution Strategy~\citep{hansen2016cma} with population size $\lambda = 4 + 3 \log d$.
\end{itemize}

All methods optimize the same tunable parameter set $\theta$ under identical bounds (defined in Section~\ref{sec:simulation}): four joint parameters for the finger, four material parameters for the tentacle in air, and three environment parameters for the underwater stress test. We will release code and configuration files upon publication (Appendix~\ref{app:hardware}).

\subsection{Ablation Study}
\label{sec:ablations}

We conduct ablation studies on the \method{} pipeline to isolate individual design choices. Each ablation modifies exactly one factor relative to the full method:
\begin{itemize}
    \item \emph{w/o Video} (text-only): Video media removed from the VLM prompt. Only scalar error metrics, parameter values, and bounds are provided. All other factors (history, CoT, control tuning) remain unchanged.
    \item \emph{w/o Chain-of-Thought}: The ``\emph{Let's think step by step}'' instruction~\citep{wei2022chain} is removed from the system prompt. Video and all other factors remain.
    \item \emph{w/o History} (no in-context learning): Previous iterations' parameter-error pairs are omitted from the prompt~\citep{brown2020language}. The VLM sees only the current state.
    \item \emph{w/o Active Learning} (fixed control): Control parameters are locked to training values. The VLM can only recommend physics parameter changes.
\end{itemize}
Ablation results are presented in Section~\ref{sec:ablation_results}.

\section{Results}
\label{sec:results}

We lead with holdout generalization as the primary success metric (evaluation protocol in Section~\ref{sec:eval_protocol}). Concretely, we evaluate each method's best training iteration on unseen control profiles (H1--H4).
Unless stated otherwise, we report mean $\pm$ std over $N{=}3$ training seeds. Each seed value aggregates $4$ holdouts $\times$ $3$ hardware repeats ($12$ datapoints). Reported uncertainty therefore reflects algorithmic sensitivity, not per-trial hardware noise.

\subsection{\simtoreal{} Holdout Generalization}

Table~\ref{tab:holdout_all} summarizes \simtoreal{} holdout generalization across all three experimental settings. The rightmost column reports the average rank. For each setting, methods are ranked 1--6 by mean holdout error, and the rank is averaged across all three settings.

\emph{Finger.}
\method{} achieves the lowest mean holdout error (10.9~px), a 10\% reduction vs.\ the next-best baseline Golden-CD (12.1~px).

\emph{Tentacle (air).}
Multiple methods achieve similar holdout performance. \method{} (53.0~px) is within 2\% of the best baseline CMA-ES (52.1~px), while providing interpretable rationales that aid debugging (see Section~\ref{sec:discussion}).

\emph{Tentacle (water).}
Under the shared-body protocol, all methods start from body parameters discovered by \method{} during in-air calibration and tune only environment parameters. All methods cluster between 71.7--80.7~px with overlapping uncertainty intervals, despite baselines receiving a head start from parameters they did not discover. BO achieves the lowest mean (71.7~px), followed by \method{} (73.3~px). The narrow spread (range $<$10~px) indicates that \emph{optimizer choice matters less than model class}. The simplified drag model cannot fully capture fluid-structure interaction, so tuning saturates quickly regardless of optimizer.

\emph{Per-seed analysis.}
Notably, \method{} achieves the best single-seed result across all three domains (4.9~px finger, 48.8~px tentacle, 71.1~px water; Table~\ref{tab:holdout_all} parenthesized values), showing that when the optimizer ``gets it right,'' VLM reasoning outperforms all baselines.
This comes with moderate cross-seed variance (std=6.3~px on finger), whereas Golden-CD shows remarkable consistency (std=0.2~px).
This variance reflects a trade-off between exploration and consistency.
\method{}'s stochasticity arises from VLM sampling (temperature 1.0), which enables the bold cross-parameter corrections shown in Section~\ref{sec:convergence} but also produces occasional poor seeds (e.g., Seed~3 at 19.5~px on finger; Appendix~\ref{app:per_seed}).
Golden-CD's low variance stems from its deterministic coordinate-descent strategy, which reliably finds good local optima but cannot make large joint parameter changes.
The gap between \method{}'s best seed (4.9~px) and Golden-CD's best (11.9~px) suggests that reducing VLM variance through ensembling or temperature annealing could yield further gains.

\begin{table*}[t]
    \centering
    \caption{\simtoreal{} holdout generalization across experimental settings. We report mean $\pm$ std over $N=3$ training seeds, with the best single-seed result in parentheses. Each seed value is the mean over 4 holdouts $\times$ 3 hardware repeats (12 datapoints). Best mean in \textbf{bold}, second-best \underline{underlined}. The water column uses a shared-body protocol (environment parameters only) and is included as a stress test. All methods perform similarly due to model-class limitations.}
    \label{tab:holdout_all}
    \small
    \setlength{\tabcolsep}{8pt}
    \begin{tabular}{lcccc}
        \toprule
        \textbf{Method} & \textbf{Finger (px) $\downarrow$} & \textbf{Tentacle--Air (px) $\downarrow$} & \textbf{Tentacle--Water (px) $\downarrow$}$^\dagger$ & \textbf{Avg.\ Rank $\downarrow$} \\
        \midrule
        Random & 27.8 $\pm$ 17.2\,(12.1) & 54.3 $\pm$ 1.9\,(51.6) & 76.1 $\pm$ 3.8\,(71.2) & 4.3 \\
        Nelder-Mead & 17.2 $\pm$ 3.5\,(14.2) & 57.3 $\pm$ 4.2\,(54.3) & 80.7 $\pm$ 2.9\,(77.3) & 5.3 \\
        Golden-CD & \underline{12.1 $\pm$ 0.2}\,(11.9) & 53.4 $\pm$ 4.8\,(49.5) & 77.8 $\pm$ 3.7\,(73.2) & 3.3 \\
        BO & 16.1 $\pm$ 2.4\,(14.4) & 62.4 $\pm$ 10.4\,(54.5) & \textbf{71.7 $\pm$ 0.1}\,(71.5) & 3.7 \\
        CMA-ES & 15.3 $\pm$ 4.9\,(11.8) & \textbf{52.1 $\pm$ 2.9}\,(49.9) & 77.6 $\pm$ 3.4\,(72.9) & \underline{2.7} \\
        \rowcolor{black!8} \textbf{\method{} (Ours)} & \textbf{10.9 $\pm$ 6.3}\,(\textbf{4.9}) & \underline{53.0 $\pm$ 5.3}\,(\textbf{48.8}) & \underline{73.3 $\pm$ 3.1}\,(\textbf{71.1}) & \textbf{1.7} \\
        \bottomrule
        \multicolumn{5}{l}{\footnotesize $^\dagger$Stress test: environment-only tuning with shared body parameters. Narrow spread indicates model-class bottleneck.}
    \end{tabular}
\end{table*}

\subsection{Training Convergence}
\label{sec:convergence}

Training error (MAE) for \method{} plateaus within 5 iterations on both platforms, matching or exceeding the rate of black-box baselines (Figure~\ref{fig:convergence} in Appendix~\ref{app:per_seed}).
Note that objective-space convergence does not guarantee parameter convergence. We verify that parameters themselves converge toward ground truth in the \simtosim{} setting (Figure~\ref{fig:param_traj_distance}).
The VLM actively varies motor amplitudes across iterations to elicit informative motions, maintaining substantial amplitude throughout optimization without collapsing to degenerate low-motion solutions (Appendix~\ref{app:amplitude}).

To illustrate how VLM reasoning drives this convergence, consider a representative finger \simtoreal{} iteration.
Given paired sim-real videos, the VLM diagnoses: ``\emph{The simulation finger is significantly slower and more heavily damped than the real hardware. The simulated motion appears sluggish.}''
Based on this visual observation, it recommends reducing friction (143$\rightarrow$25) and damping (190$\rightarrow$30), producing a 59\% error reduction (50.1$\rightarrow$20.7~px) in a single step.
This diagnostic chain, connecting qualitative visual discrepancies to targeted parameter corrections, is the mechanism underlying \method{}'s convergence.
By contrast, black-box optimizers can achieve large error reductions but cannot explain which physical mismatch motivated the change.

In early iterations, bold corrections based on gross visual mismatches (motion too slow, wrong amplitude) drive rapid error reduction.
Near convergence, recommendations become more conservative and less reliable as remaining discrepancies grow subtler.
This is why we track the best parameters seen across all iterations (Algorithm~\ref{alg:vid2sid}), allowing the pipeline to benefit from early breakthroughs without being degraded by later missteps (additional reasoning examples, including a failure case, in Appendix~\ref{app:vlm_examples}).

Figure~\ref{fig:finger_qualitative} shows qualitative sim-real alignment for the finger after calibration. The calibrated simulation closely tracks both the amplitude and phase of the real finger's motion. Notably, the residual is not a constant offset but varies with the motion cycle, suggesting that remaining error stems from unmodeled nonlinearities (e.g., cable elasticity, stick-slip friction) rather than a simple parameter bias.

\begin{figure}[h]
    \centering
    \includegraphics[width=\columnwidth]{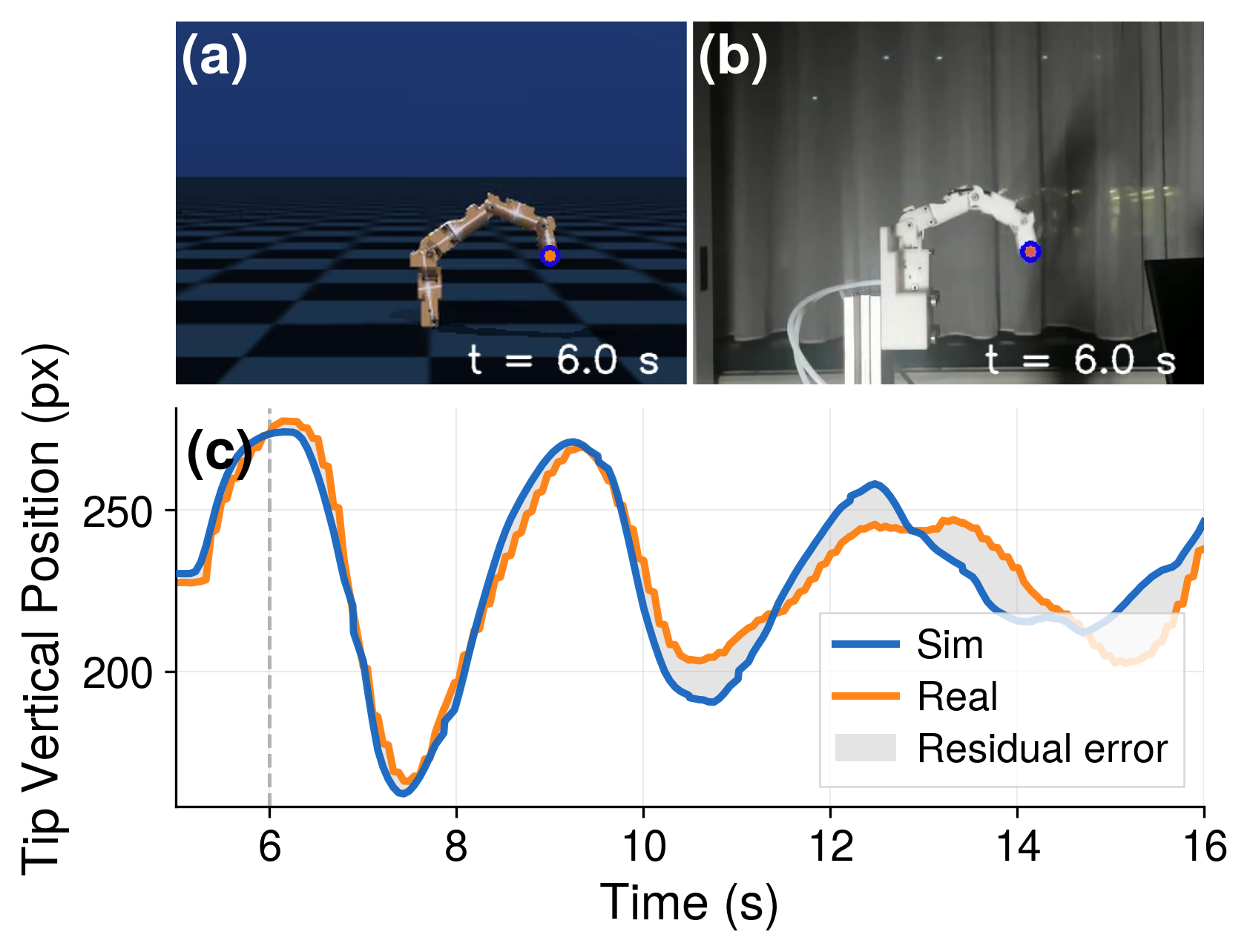}
    \caption{Finger qualitative alignment after \method{} calibration. (a)~Simulated finger at $t{=}6$\,s. (b)~Corresponding real video frame with tracked marker. (c)~Tip vertical position over time: calibrated simulation (blue) closely tracks real finger motion (orange). The narrow residual (shaded) confirms that \method{} captures both the amplitude and timing of real finger dynamics.}
    \label{fig:finger_qualitative}
\end{figure}

\subsection{Ablation Study}
\label{sec:ablation_results}

\begin{figure}[h]
    \centering
    \includegraphics[width=0.95\columnwidth]{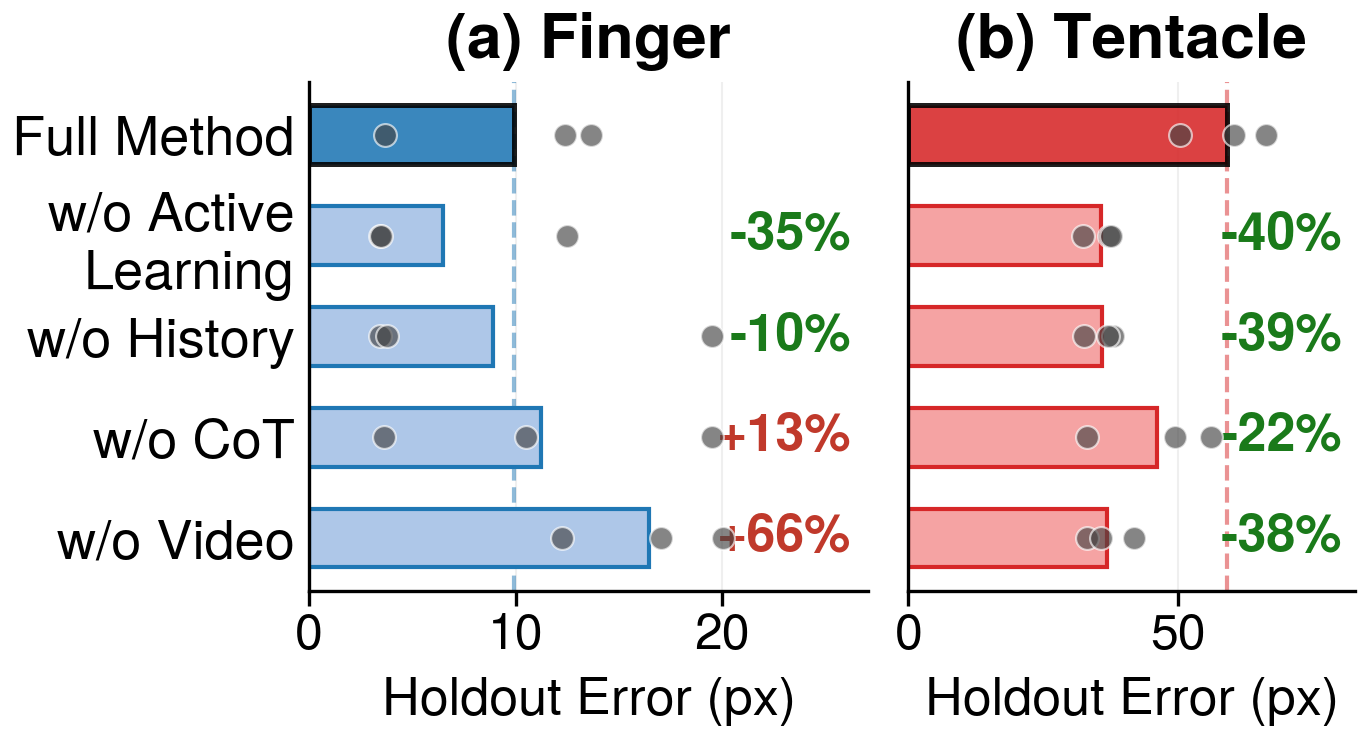}
    \caption{Ablation study on (a)~finger and (b)~tentacle. Each bar removes one component. Dots show individual seeds. Removing video increases error by 66\% on the finger but decreases it by 38\% on the tentacle, indicating that prompt design is domain-dependent.}
    \label{fig:ablations}
\end{figure}

Figure~\ref{fig:ablations} summarizes ablation results on both platforms. Each ablation modifies exactly one factor relative to the full \method{} configuration (see Section~\ref{sec:ablations} for definitions).

\emph{Finger ablations.}
On the finger, removing video input increases holdout error by 66\%, confirming that VLMs extract useful information from raw video beyond scalar metrics.
Disabling chain-of-thought prompting~\citep{wei2022chain} increases error by 13\%.
Interestingly, removing iteration history (in-context learning~\citep{brown2020language}) reduces error by 10\%, suggesting that history can anchor the VLM to early (possibly suboptimal) iterations, encouraging incremental refinements rather than bold exploration.
Fixing control inputs (disabling active learning) reduces error by 35\%, likely because varying the control signal between iterations introduces a confound. The VLM must simultaneously learn physics parameters and adapt to changing observations, making iteration-to-iteration comparison harder.
In summary, video input is the most impactful component ($+$66\%), while CoT provides a modest benefit ($+$13\%) and history and active learning slightly hurt, suggesting that the VLM performs best when given rich visual input but minimal confounds from changing observations.

\emph{Tentacle ablations.}
On the tentacle, removing video, history, or active learning each reduces holdout error by 38--40\% relative to the full configuration, while removing CoT reduces it by 22\%.
We attribute this to a fundamental difference in \emph{parameter observability} between the two platforms.
Finger parameters (damping, friction) produce visually distinctive signatures (oscillation speed, settling time, overshoot) that the VLM can reliably perceive and reason about from video.
Tentacle material properties (Young's modulus, Poisson ratio) produce subtler deformation effects that are obscured by SAM3 segmentation noise ($\sim$3--5~px centerline jitter), and the chaotic tip dynamics make frame-level reasoning unreliable.
In this regime, video and history introduce more noise than signal. History anchors the VLM to early mistakes, and varying the control signal across iterations further obscures parameter effects by changing the observation conditions simultaneously.
These results suggest that prompt design is domain-dependent and practitioners should run a small ablation when deploying \method{} on new platforms (see Section~\ref{sec:discussion}).

\subsection{\simtosim{} Validation}

\begin{figure}[!t]
    \centering
    \includegraphics[width=\columnwidth]{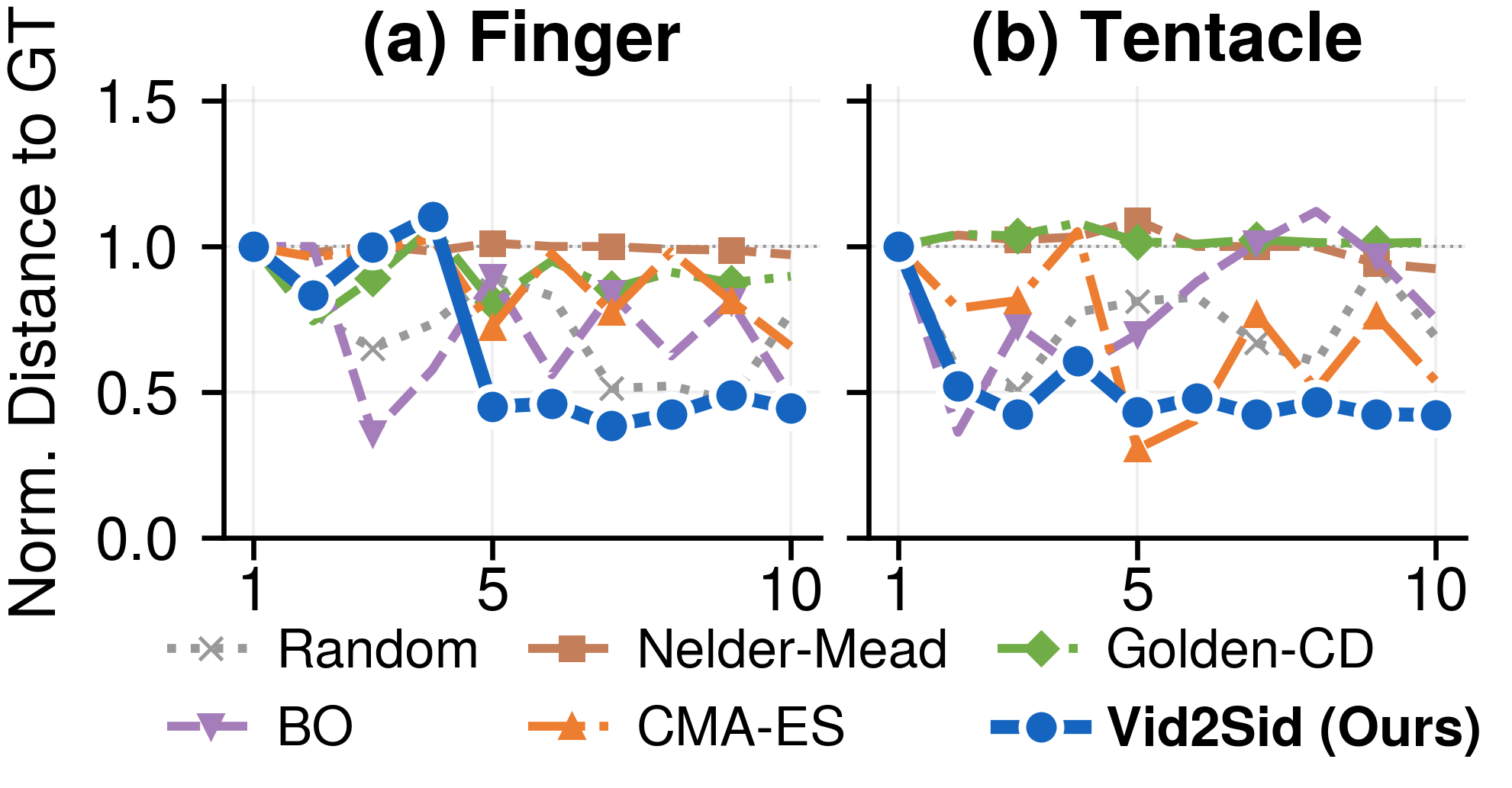}
    \caption{Normalized distance to ground truth over iterations (\simtosim{}, best seed per method). \method{} consistently converges toward the true parameters on both platforms, while baselines oscillate or settle at compensating values far from ground truth. This confirms that VLM reasoning recovers physically meaningful parameters rather than exploiting compensating combinations.}
    \label{fig:param_traj_distance}
\end{figure}

We validate the pipeline in a controlled \simtosim{} setting where the ``real'' data is also simulated with known ground truth parameters.
Table~\ref{tab:sim2sim} shows holdout errors for both platforms (finger error is reported in mm because both ``sim'' and ``real'' operate in world coordinates, unlike \simtoreal{} which compares in pixel space).
\method{} achieves the best performance on both finger (7.13~mm, a 27\% improvement over BO) and tentacle (8.91~px, an 11\% improvement over BO/CMA-ES), demonstrating robust parameter recovery when model misspecification is eliminated.
Appendix~\ref{app:param_recovery} shows that \method{} also recovers parameters closest to ground truth (mean relative error 8.7\% on finger, 12.4\% on tentacle vs.\ 28--98\% for baselines), whereas black-box methods often find compensating parameter combinations far from the true values.
Figure~\ref{fig:param_traj_distance} visualizes this directly. \method{} steadily decreases its distance to ground truth over iterations, while baselines oscillate or stagnate in parameter space (per-parameter trajectories in Appendix~\ref{app:param_trajectories}).
This result is particularly informative in light of the water stress test. \method{} excels when the simulator can faithfully represent the real dynamics, confirming that residual \simtoreal{} gaps in the water setting are attributable to model class limitations rather than optimizer deficiency.

\begin{table}[!tb]
    \centering
    \caption{\simtosim{} holdout generalization (mean $\pm$ std over $N=12$ datapoints: 3 seeds $\times$ 4 holdouts). Finger error is in mm (2D tip position); tentacle error is in px (2D centerline). Without model misspecification, \method{} recovers parameters most consistently across both platforms, suggesting that residual \simtoreal{} error is dominated by perception and model-class limits.}
    \label{tab:sim2sim}
    \small
    \begin{tabular}{lcc}
        \toprule
        \textbf{Method} & \textbf{Finger (mm) $\downarrow$} & \textbf{Tentacle (px) $\downarrow$} \\
        \midrule
        Random & 15.67 $\pm$ 13.80 & 10.63 $\pm$ 9.59 \\
        Nelder-Mead & 46.41 $\pm$ 8.77 & 11.65 $\pm$ 11.05 \\
        Golden-CD & 16.36 $\pm$ 17.79 & 11.44 $\pm$ 12.44 \\
        BO & 9.70 $\pm$ 7.49 & 10.02 $\pm$ 11.03 \\
        CMA-ES & 23.77 $\pm$ 19.23 & 10.02 $\pm$ 7.15 \\
        \rowcolor{black!8} \textbf{\method{} (Ours)} & \textbf{7.13 $\pm$ 2.56} & \textbf{8.91 $\pm$ 9.52} \\
        \bottomrule
    \end{tabular}
\end{table}

\section{Discussion}
\label{sec:discussion}

Our results reveal three calibration regimes determined by observation quality and simulator expressiveness.
(1)~When perception is clean and the simulator is expressive (finger), VLM video reasoning drives strong gains: 10\% holdout improvement and a 66\% ablation benefit from video input (Figure~\ref{fig:ablations}).
(2)~When perception is noisy (tentacle in air), \method{} remains competitive but video input introduces noise from segmentation artifacts ($\sim$3--5~px centerline jitter), so text-only mode is preferable for its interpretability benefits.
(3)~When model misspecification dominates (water), all optimizers converge to similar error ($<$10~px spread), confirming that model class bounds performance regardless of optimizer.

\textbf{Practical guideline.}
Deploy \method{} with video when perception is clean and the simulator is expressive. Under noisy segmentation, use text-only mode or fall back to a black-box optimizer.

Beyond accuracy, interpretability aids debugging by surfacing falsifiable hypotheses (e.g., missing forces, incorrect damping) that black-box optimizers cannot provide (reasoning examples in Appendix~\ref{app:vlm_examples}).
The VLM's self-reported confidence scores are also well-calibrated: recommendations succeed 89\% of the time at confidence $\geq$0.9 (Appendix~\ref{app:confidence}).

\subsection{Limitations and Failure Modes}

We identify four failure modes and corresponding safeguards:
(1)~\textbf{Segmentation failure} from glare or occlusion, mitigated by adding point prompts to SAM3.
(2)~\textbf{Timing drift} from recording latency, addressed via cross-correlation alignment (max 1\,s lag).
(3)~\textbf{Degenerate control updates} where the VLM reduces motion amplitude to trivially lower error, prevented by minimum amplitude bounds and holdout evaluation.
(4)~\textbf{Invalid VLM recommendations} with implausible values, clamped to valid bounds.

\textbf{Computational cost.}
\method{} adds $\sim$11\,s of VLM inference per iteration ($\sim$28\% overhead, \$0.14 per 10-iteration run), keeping total calibration under 10 minutes (Appendix~\ref{app:wallclock}). This is acceptable for offline use but precludes real-time adaptation.

\textbf{Scope limitations.}
Our pipeline assumes a single fixed camera with a static background and does not handle occlusions, lighting changes, or multi-robot scenarios.
VLM recommendations are not guaranteed to be globally optimal. We observed occasional convergence to local minima that black-box methods avoided, and extremely poor initializations (e.g., Young's modulus off by $100\times$) caused all methods to fail.
Additionally, we evaluated only Google Gemini 2.5 Pro. As VLMs with stronger physical reasoning emerge, video input may become beneficial on a wider range of platforms (see Appendices~\ref{app:comparison} and~\ref{app:deployment} for a detailed comparison with black-box optimizers and deployment guidance).

\section{Conclusion}
\label{sec:conclusion}

We introduce \method{}, a camera-only \simtoreal{} calibration pipeline that turns sim-real videos into physics parameter updates via foundation-model tracking and VLM reasoning.
Across a tendon-driven finger and a soft tentacle, \method{} matches or outperforms black-box baselines on holdout generalization within 10 iterations, requires no optimizer hyperparameters, and uniquely provides natural-language rationales that make calibration \emph{debuggable}.
Our experiments reveal three calibration regimes: VLM reasoning excels when perception is clean and the simulator is expressive (finger), provides interpretability benefits when perception is noisy (tentacle in air), and performs comparably to all methods when model misspecification dominates (tentacle in water).

\textbf{Future directions.}
Three extensions would broaden \method{}'s applicability: (1)~distilling perception and VLM inference for real-time online calibration during deployment, (2)~extending the observation model to multi-view and multi-robot settings with occlusions and dynamic backgrounds, and (3)~fine-tuning foundation models specifically for physics reasoning to reduce domain-dependent prompt sensitivity.
We will release code and experiment configurations to support future work.

\bibliographystyle{plainnat}
\bibliography{references}

\clearpage
\appendices

\onecolumn
\begin{center}
{\LARGE\bfseries Table of Contents}
\end{center}
\vspace{0.3em}
\noindent\rule{\textwidth}{0.5pt}

\apptocmain{A~~Experimental Setup}{app:hardware}
\apptocsub{A.1~~Finger Platform}{app:finger_hw}
\apptocsub{A.2~~Tentacle Platform}{app:tentacle_hw}
\apptocsub{A.3~~Compute Resources}{app:compute}
\apptocsub{A.4~~Wall-Clock Timing}{app:wallclock}

\apptocmain{B~~Simulation Model Details}{app:sim_models}
\apptocsub{B.1~~Finger Model (MuJoCo)}{app:finger_model}
\apptocsub{B.2~~Tentacle Model (PyElastica)}{app:tentacle_model}

\apptocmain{C~~Parameter Bounds}{app:bounds}
\apptocsub{C.1~~Finger Platform (MuJoCo)}{app:bounds_finger_sub}
\apptocsub{C.2~~Tentacle Platform (PyElastica)}{app:bounds_tentacle_sub}

\apptocmain{D~~Evaluation Protocol}{app:eval_protocol}

\apptocmain{E~~Baseline Implementation}{app:baselines}
\apptocsub{E.1~~Bayesian Optimization (BO)}{app:bo}
\apptocsub{E.2~~CMA-ES}{app:cmaes}
\apptocsub{E.3~~Nelder-Mead}{app:nelder}
\apptocsub{E.4~~Golden-CD}{app:golden_cd}
\apptocsub{E.5~~Random Search}{app:random}

\apptocmain{F~~VLM Pipeline}{app:prompts}
\apptocsub{F.1~~Prompt Templates}{app:prompt_templates}
\apptocsub{F.2~~Cost and Latency Analysis}{app:vlm_cost}
\apptocsub{F.3~~VLM Inference Settings}{app:vlm_settings}
\apptocsub{F.4~~Confidence Calibration}{app:confidence}
\apptocsub{F.5~~VLM Reasoning Examples}{app:vlm_examples}

\apptocmain{G~~SAM3 Perception}{app:sam3}
\apptocsub{G.1~~Segmentation Reliability}{app:sam3_reliability}
\apptocsub{G.2~~Text Prompts}{app:sam3_prompts}
\apptocsub{G.3~~Fallback Strategy}{app:sam3_fallback}

\apptocmain{H~~Extended Results}{app:per_seed}
\apptocsub{H.1~~Training Convergence}{app:training_convergence}
\apptocsub{H.2~~Per-Seed Training Error}{app:perseed_training}
\apptocsub{H.3~~Per-Seed Holdout Error}{app:perseed_holdout}
\apptocsub{H.4~~Per-Holdout Error Breakdown}{app:per_holdout_breakdown}
\apptocsub{H.5~~Ablation Per-Seed Results}{app:ablation_seeds}
\apptocsub{H.6~~Active Learning Amplitude Trajectories}{app:amplitude}
\apptocsub{H.7~~\simtosim{} Parameter Recovery}{app:param_recovery}
\apptocsub{H.8~~Parameter Trajectories}{app:param_trajectories}

\apptocmain{I~~Comparison with Black-Box Optimization}{app:comparison}

\apptocmain{J~~Deployment Guidelines}{app:deployment}

\twocolumn

\section{Experimental Setup}
\label{app:hardware}

\subsection{Finger Platform}
\label{app:finger_hw}

The finger is 3D-printed from PLA and mounted on aluminum extrusion.
Two Dynamixel XC330-T288-T motors drive it through Bowden cables. The MCP joint is antagonistically driven for flexion and extension, while the PIP and DIP joints are coupled and returned by an elastic string.
A visual marker on the fingertip enables tracking.
Video is captured with a Logitech BRIO at 1920$\times$1080 and 30\,fps.

\subsection{Tentacle Platform}
\label{app:tentacle_hw}

The tentacle is mold-cast from DragonSkin 10 Slow silicone and attached to a Dynamixel XW430-T200R motor.
The entire assembly is fixed to the side of a water tank, with the tentacle submerged for underwater experiments.
For in-air experiments, video is captured with a Logitech BRIO at 640$\times$480 and 25\,fps.
For underwater experiments, video is captured with a GoPro Hero 6 at 1920$\times$1080 and 120\,fps (Superview FOV).

\subsection{Compute Resources}
\label{app:compute}

Table~\ref{tab:compute} lists the hardware and software used for all experiments.

\begin{table}[!htb]
    \centering
    \caption{Compute resources.}
    \label{tab:compute}
    \small
    \begin{tabular}{ll}
        \toprule
        \textbf{Component} & \textbf{Specification} \\
        \midrule
        CPU & Intel Core i7-10700K \\
        RAM & 32 GB DDR4 \\
        GPU (SAM3) & NVIDIA A100 40GB \\
        VLM & Google Gemini 2.5 Pro \\
        OS & Ubuntu 22.04 LTS \\
        \bottomrule
    \end{tabular}
\end{table}

\subsection{Wall-Clock Timing}
\label{app:wallclock}

Table~\ref{tab:wallclock} breaks down per-iteration wall-clock time for the tentacle platform. Hardware capture, simulation, and SAM3 segmentation are constant across all methods; the only variable is the optimizer step. \method{} adds $\sim$11\,s of VLM API latency per iteration, while black-box optimizers finish in under 1\,s.

\begin{table}[!htb]
    \centering
    \caption{Per-iteration wall-clock time (tentacle platform). Shared components (hardware capture: 15\,s, simulation: 8\,s, SAM3: 12\,s) total 35\,s for all methods. \method{} adds 11\,s VLM inference (28\% overhead), keeping total calibration under 10 minutes.}
    \label{tab:wallclock}
    \small
    \begin{tabular}{lcc}
        \toprule
        & \textbf{\method{}} & \textbf{Baselines} \\
        \midrule
        Shared components & 35\,s & 35\,s \\
        Optimizer/VLM & 11\,s & $<$1\,s \\
        \midrule
        \textbf{Total per iteration} & \textbf{46\,s} & \textbf{36\,s} \\
        \textbf{10-iteration calibration} & 7.7\,min & 6.0\,min \\
        \bottomrule
    \end{tabular}
\end{table}

The 28\% per-iteration overhead translates to $\sim$1.7 minutes additional wall-clock time for a full 10-iteration calibration run. This overhead is offset by eliminating the hyperparameter tuning that black-box methods typically require offline.

\section{Simulation Model Details}
\label{app:sim_models}

\subsection{Finger Model (MuJoCo)}
\label{app:finger_model}

The MuJoCo finger model was generated from a SolidWorks CAD assembly using the following pipeline:
\begin{enumerate}
    \item Design finger assembly in SolidWorks with proper joint definitions
    \item Export to URDF using \textbf{sw\_urdf\_exporter}~\cite{sw_urdf_exporter}, a ROS plugin that converts SolidWorks assemblies to URDF format with automatic mesh export
    \item Convert URDF to MJCF (MuJoCo XML) using the \textbf{urdf2mjcf} library~\cite{urdf2mjcf}
    \item Manually tune default physics parameters (damping, friction) to approximate real behavior
\end{enumerate}

The baseline profile uses hand-tuned default values chosen to produce qualitatively similar motion.
For experiments, initial parameters are randomly sampled from bounds for each training seed (see Section~\ref{sec:exp_sim2real}).

\subsection{Tentacle Model (PyElastica)}
\label{app:tentacle_model}

The PyElastica tentacle model uses a Cosserat rod formulation with 60 discretization elements along the rod length. Actuation is applied as a muscle torque at the base element. The model includes perpendicular and tangential fluid drag forces and Rayleigh damping for numerical stability.

The baseline profile uses material properties estimated from manufacturer datasheets for DragonSkin 10 Slow silicone (Shore 10A hardness, $\rho \approx 1050$ kg/m$^3$, $E \approx 0.2$ MPa).
For experiments, initial parameters are randomly sampled from bounds for each training seed.

\section{Parameter Bounds}
\label{app:bounds}

\subsection{Finger Platform (MuJoCo)}
\label{app:bounds_finger_sub}

Table~\ref{tab:bounds_finger} lists the 6 parameters tuned for the finger, spanning physics properties (friction, damping, inertia, density) and control inputs (joint amplitudes). Nominal values reflect hand-tuned baseline estimates; experiments initialize by sampling uniformly within bounds.

\begin{table*}[t]
    \centering
    \caption{Parameter bounds for the finger platform.}
    \label{tab:bounds_finger}
    \small
    \begin{tabular}{lcccl}
        \toprule
        \textbf{Parameter} & \textbf{Min} & \textbf{Max} & \textbf{Nominal} & \textbf{Description} \\
        \midrule
        Friction loss & 0 & 150 & 50 & Joint dry friction torque (Nm) \\
        Damping & 10 & 200 & 100 & Velocity-dependent viscous damping \\
        Armature & 0.1 & 5.0 & 1.0 & Rotor inertia added to each joint \\
        Link density & 1.0 & 20.0 & 5.0 & Density of body geometries (kg/m$^3$) \\
        MCP amplitude & 0 & 60 & 10 & MCP joint oscillation amplitude (deg) \\
        PIP amplitude & 0 & 60 & 10 & PIP/DIP joint oscillation amplitude (deg) \\
        \bottomrule
    \end{tabular}
\end{table*}

\subsection{Tentacle Platform (PyElastica)}
\label{app:bounds_tentacle_sub}

Table~\ref{tab:bounds_tentacle} shows the body, environment, and control parameter bounds for the tentacle platform. Nominal values are estimated from manufacturer datasheets; experiments initialize by sampling uniformly within bounds. During in-air calibration, only body parameters are tuned; environment parameters become active in the underwater setting. Table~\ref{tab:bounds_water} lists the environment bounds used for the underwater stress test, with body parameters frozen from the in-air experiment.

\begin{table*}[t]
    \centering
    \caption{Parameter bounds for the tentacle platform (in-air calibration).}
    \label{tab:bounds_tentacle}
    \small
    \begin{tabular}{lcccl}
        \toprule
        \textbf{Parameter} & \textbf{Min} & \textbf{Max} & \textbf{Nominal} & \textbf{Description} \\
        \midrule
        \multicolumn{5}{l}{\emph{Body parameters}} \\
        Young's modulus & $5\times10^3$ & $5\times10^6$ & $2\times10^5$ & Rod stiffness (Pa) \\
        Rod density & 500 & 5000 & 1050 & Material density (kg/m$^3$) \\
        Poisson's ratio & 0.2 & 0.5 & 0.5 & Lateral-to-axial strain ratio \\
        Damping constant & 0 & 100 & 3.0 & Linear (Rayleigh) damping coefficient \\
        \midrule
        \multicolumn{5}{l}{\emph{Environment parameters}} \\
        Perpendicular drag & 0 & 50 & 1.0 & Perpendicular fluid drag coefficient \\
        Tangential drag & 0 & 50 & 0.1 & Tangential fluid drag coefficient \\
        \midrule
        \multicolumn{5}{l}{\emph{Control parameters}} \\
        Motor amplitude & 0.2 & 1.0 & 1.0 & Oscillation amplitude (rad), tuned via active learning \\
        Motor frequency & -- & -- & 0.15 & Fixed during training; holdouts use 0.15 and 0.5\,Hz \\
        \bottomrule
    \end{tabular}
\end{table*}

\begin{table*}[t]
    \centering
    \caption{Hydrodynamic environment parameter bounds for the underwater tentacle stress test. Body parameters are frozen from the in-air experiment.}
    \label{tab:bounds_water}
    \small
    \begin{tabular}{lcccl}
        \toprule
        \textbf{Parameter} & \textbf{Min} & \textbf{Max} & \textbf{Nominal} & \textbf{Description} \\
        \midrule
        Fluid density & 0 & 2000 & 1000 & Surrounding fluid density (kg/m$^3$) \\
        Perpendicular drag & 0 & 100 & 10 & Perpendicular fluid drag coefficient \\
        Tangential drag & 0 & 100 & 10 & Tangential fluid drag coefficient \\
        \bottomrule
    \end{tabular}
\end{table*}

\section{Evaluation Protocol}
\label{app:eval_protocol}

Our evaluation captures two distinct sources of variance:

\textbf{Algorithmic variance} (training seeds). We run each method with 3 different random seeds. The seed controls initial parameter sampling for all methods, the stochastic optimizer state for CMA-ES and BO, and the VLM sampling for \method{}.

\textbf{Hardware variance} (execution repeatability). For each trained parameter configuration, we execute 3 independent hardware trials to capture motor variability, material hysteresis, temperature effects, and camera timing jitter.

\textbf{Reporting convention.}
Main results (Table~\ref{tab:holdout_all}) report mean $\pm$ std over the $N=3$ \emph{training seeds}.
For each training seed, the holdout error is the mean over 4 holdouts $\times$ 3 hardware repeats (12 datapoints), giving 36 holdout datapoints per method and domain.
The reported uncertainty therefore reflects sensitivity to random initialization and optimizer stochasticity, not per-trial hardware noise.
When needed, we additionally report raw hardware-repeat variability in supplementary analyses.

\textbf{Isolating effects.}
To isolate algorithmic vs.\ hardware effects, we select the best-performing iteration from each training seed independently, then evaluate each on all 4 holdout controls with 3 hardware repeats.
This ensures that hardware variance does not contaminate training seed selection.

\textbf{Holdout profiles (H1--H4).}
Table~\ref{tab:holdout_profiles} specifies the control parameters for each holdout profile. Both platforms use a $2{\times}2$ factorial design spanning corners of the control space.

\begin{table}[!htb]
    \centering
    \caption{Holdout control profiles for each platform. Finger profiles vary joint amplitudes with fixed frequencies ($f_0{=}0.3$\,Hz, $f_1{=}0.35$\,Hz, phase offset$=45^\circ$). Tentacle profiles vary amplitude and frequency.}
    \label{tab:holdout_profiles}
    \small
    \begin{tabular}{lcc}
        \toprule
        & \textbf{Finger} & \textbf{Tentacle} \\
        & ($A_0$, $A_1$) & ($A$, $f$) \\
        \midrule
        H1 & $50^\circ$, $50^\circ$ & 0.9\,rad, 0.5\,Hz \\
        H2 & $50^\circ$, $8^\circ$ & 0.9\,rad, 0.15\,Hz \\
        H3 & $8^\circ$, $50^\circ$ & 0.3\,rad, 0.5\,Hz \\
        H4 & $8^\circ$, $8^\circ$ & 0.3\,rad, 0.15\,Hz \\
        \bottomrule
    \end{tabular}
\end{table}

\section{Baseline Implementation}
\label{app:baselines}

All baselines follow the same evaluation protocol. Each method runs for 10 iterations per seed across 3 training seeds, and the best iteration is selected for holdout evaluation.

\subsection{Bayesian Optimization (BO)}
\label{app:bo}

We use \texttt{scikit-learn}'s Gaussian Process implementation with:
\begin{itemize}
    \item Surrogate model: Gaussian Process with Mat\'ern 5/2 kernel
    \item Acquisition function: Expected Improvement (EI)
    \item Initial points: 3 random samples before GP fitting
    \item Kernel hyperparameters: optimized via marginal likelihood
    \item Bounds: normalized to $[0, 1]$ per dimension
\end{itemize}

\subsection{CMA-ES}
\label{app:cmaes}

We use the \texttt{cma} Python package with:
\begin{itemize}
    \item Initial $\sigma$: 0.3 (normalized parameter space)
    \item Population size: $4 + \lfloor 3 \ln(d{+}1) \rfloor$ where $d$ = dimension
    \item Bounds handling: reflection at boundaries
    \item Termination: 10 iterations (not function evaluations)
\end{itemize}

Because CMA-ES is population-based, it evaluates multiple candidates per iteration. We count each full population evaluation as one ``iteration'' for fair comparison, so CMA-ES uses more total function evaluations than other methods.
This accounting favors CMA-ES, making our comparisons against \method{} conservative.

\subsection{Nelder-Mead}
\label{app:nelder}

We use \texttt{scipy.optimize.minimize} with:
\begin{itemize}
    \item Initial simplex: scipy default construction from the initial point
    \item Reflection/expansion/contraction coefficients: default (1, 2, 0.5)
    \item Bounds handling: parameter values clamped to the feasible region at each evaluation
    \item Termination: 10 function evaluations
\end{itemize}

\subsection{Golden Section / Coordinate Descent (Golden-CD)}
\label{app:golden_cd}

This method performs sequential 1D optimization along each parameter axis using golden section search with tolerance-based convergence. The method cycles through all parameters, distributing the total budget of 10 evaluations across dimensions.
Its deterministic nature yields the most consistent finger performance (std=0.2~px across seeds).

\subsection{Random Search}
\label{app:random}

We draw 10 independent samples uniformly within parameter bounds per seed and select the best for holdout evaluation, with no sequential refinement.
Despite its simplicity, Random achieved competitive performance on the tentacle, suggesting that the objective landscape contains multiple good local optima.

\section{VLM Pipeline}
\label{app:prompts}

\subsection{Prompt Templates}
\label{app:prompt_templates}

At each iteration the VLM receives a structured prompt containing the system instruction, current parameter values, error metrics, parameter bounds, iteration history, and two videos (simulation and real).
Simplified versions of the prompt components are shown below. Full templates are available in the supplementary code.

\subsubsection{System Prompt}

The system prompt is a single instruction paragraph. The version shown is for the finger platform; the tentacle variant substitutes PyElastica-specific terms.

\noindent
\begin{promptbox}
You are an expert in robotics sim-to-real transfer and dynamic calibration. Compare the MuJoCo simulation against a real hardware capture to diagnose dynamic mismatches. Use visual cues plus the history table to reason about parameter updates. Treat amp_deg1 as a controllable probe: vary it (within bounds) to expose differences in frictionloss, damping, armature, and density, not just to cosmetically match one clip. The real hardware clip is ground truth; the simulation clip is what you are tuning.
\end{promptbox}

\subsubsection{User Prompt Structure}

Each iteration appends the current parameter profile, metrics, bounds, and history in delimited sections. A finger example is shown below.

\noindent
\begin{promptbox}
[VIDEO: simulation.mp4]
[VIDEO: real.mp4]

--- CANDIDATE PROFILE (JSON) ---
{"frictionloss": 50.0, "damping": 100.0,
 "armature": 1.0, "density": 5.0, "amp_deg1": 30.0}

--- METRICS (JSON) ---
{"mean_abs_px": 32.1}

--- PARAMETER BOUNDS ---
frictionloss: [0, 150]
damping: [10, 200]
armature: [0.1, 5.0]
density: [1.0, 20.0]
amp_deg1: [0, 60]

--- PARAMETER HISTORY ---
Iter | frictionloss | damping | error_px
  1  |    30.0      |  120.0  |  52.1
  2  |    50.0      |  100.0  |  45.2

--- TASK ---
1. Describe key discrepancies between sim and real.
2. Propose updated values for all parameters.

--- OUTPUT JSON SCHEMA (strict) ---
{
  "analysis": "Brief summary of mismatch",
  "parameter_recommendations": [
    {"name": "damping",
     "current_value": 100.0,
     "suggested_value": 70.0,
     "reason": "..."}
  ],
  "confidence": 0.75,
  "additional_notes": "..."
}
\end{promptbox}

\subsubsection{Example VLM Response}

\noindent
\begin{promptbox}
{"analysis": "Sim finger moves faster and settles
  more quickly. Real finger oscillates more.",
 "parameter_recommendations": [
  {"name": "damping", "current_value": 100.0,
   "suggested_value": 70.0,
   "reason": "Less damping allows more oscillation,
     matching real settling behavior."},
  {"name": "armature", "current_value": 1.0,
   "suggested_value": 1.5,
   "reason": "More inertia slows the motion to
     match the real finger."}],
 "confidence": 0.75,
 "additional_notes": "Other params unchanged."}
\end{promptbox}

\subsection{Cost and Latency Analysis}
\label{app:vlm_cost}

We use Google Gemini 2.5 Pro for VLM-based parameter recommendations. Table~\ref{tab:vlm_cost} summarizes the per-iteration token usage and costs based on Gemini API pricing~\cite{artificialanalysis2026}.

\textbf{Token estimation per iteration.}
\begin{itemize}
    \item System prompt: $\sim$200 tokens
    \item User prompt (parameters, bounds, metrics), $\sim$300 tokens
    \item Iteration history (averaged over 10 iterations), $\sim$500 tokens
    \item Videos (2 $\times$ 10\,s at 1\,fps $\times$ 258 tokens/frame), $\sim$5,200 tokens
    \item \textbf{Total input}, $\sim$6,200 tokens/iteration
    \item \textbf{Output} (JSON with analysis and recommendations), $\sim$600 tokens/iteration
\end{itemize}

\begin{table}[!htb]
    \centering
    \caption{VLM cost analysis for Gemini 2.5 Pro. A full 10-iteration calibration run costs \$0.14, negligible relative to hardware and compute time.}
    \label{tab:vlm_cost}
    \small
    \begin{tabular}{lcc}
        \toprule
        \textbf{Component} & \textbf{Tokens} & \textbf{Cost (USD)} \\
        \midrule
        Input (per iteration) & 6,200 & \$0.008 \\
        Output (per iteration) & 600 & \$0.006 \\
        \midrule
        \textbf{Per iteration} & 6,800 & \$0.014 \\
        \textbf{Per training run} (10 iter) & 68,000 & \$0.14 \\
        \textbf{Per experiment} (3 seeds) & 204,000 & \$0.42 \\
        \bottomrule
    \end{tabular}
\end{table}

\textbf{Latency breakdown.}
At Gemini 2.5 Pro's output speed of 151.1 tokens/second~\cite{artificialanalysis2026}, generating 600 output tokens takes $\sim$4.0 seconds.
Combined with input processing and network latency ($\sim$7.2 seconds), total VLM inference time is $\sim$11.2 seconds per iteration, comparable to the real-world capture time (10 seconds) and SAM3 segmentation (12.5 seconds).

\textbf{Pricing notes.}
Costs are based on Gemini 2.5 Pro standard tier pricing at \$1.25/1M input tokens and \$10.00/1M output tokens (for prompts $\leq$200k tokens).
Video tokens are estimated using Gemini's image tokenization rate of 258 tokens per frame at 1\,fps sampling.

\subsection{VLM Inference Settings}
\label{app:vlm_settings}

Table~\ref{tab:vlm_settings} lists the inference settings used for all \method{} experiments. We use the default Gemini 2.5 Pro configuration without any task-specific tuning. All settings are held fixed across seeds, domains (finger, tentacle, water), and experimental conditions (\simtosim{} and \simtoreal{}).

\begin{table}[!htb]
    \centering
    \caption{VLM inference settings. All values are Gemini 2.5 Pro defaults, held fixed across every seed, domain, and experimental condition.}
    \label{tab:vlm_settings}
    \small
    \begin{tabular}{ll}
        \toprule
        \textbf{Setting} & \textbf{Value} \\
        \midrule
        Model & Google Gemini 2.5 Pro \\
        Temperature & 1.0 (default) \\
        Top-$P$ & 0.95 (default) \\
        Top-$K$ & 64 (default) \\
        Max output tokens & 65{,}536 \\
        Max input tokens & 1{,}048{,}576 \\
        Safety filters & Disabled \\
        System instruction & Fixed (Appendix~\ref{app:prompts}) \\
        \bottomrule
    \end{tabular}
\end{table}

\noindent Using default decoding settings means that stochasticity across seeds arises solely from the VLM's sampling distribution, not from deliberate prompt or temperature variation. This simplifies reproducibility. Any practitioner with API access to Gemini 2.5 Pro can replicate our setup without hyperparameter search over inference settings.

\subsection{Confidence Calibration}
\label{app:confidence}

We analyze whether the VLM's self-reported confidence scores predict actual recommendation success. For each iteration, we extract the confidence score from the VLM output and measure whether the recommendation reduced error.

\subsubsection{Methodology}

We analyze $n=106$ VLM recommendations pooled across both platforms, both experimental conditions (\simtosim{} and \simtoreal{}), 3 seeds each, and 9 iterations per seed with valid before/after error measurements. We exclude 2 outlier cases where the initial simulation was catastrophically unstable (error $>$100~px), as these represent diagnostic recovery rather than normal optimization.

A recommendation is counted as a \emph{success} if error decreased after applying it. We measure \emph{precision at threshold $\tau$} as the success rate among recommendations with confidence $\geq \tau$.

\subsubsection{Results}

\begin{figure}[!htb]
    \centering
    \includegraphics[width=0.85\columnwidth]{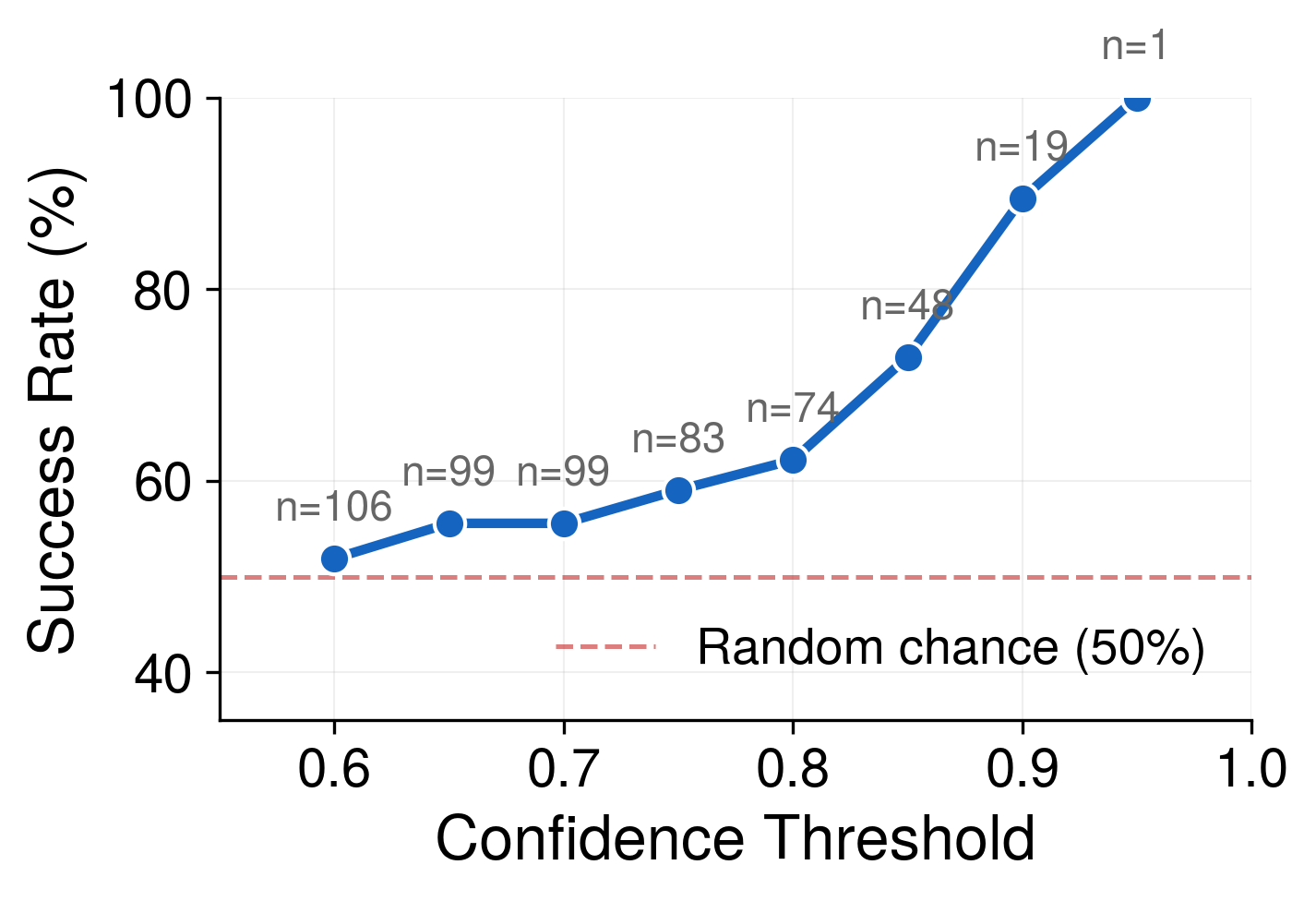}
    \caption{VLM recommendation success rate by confidence threshold. Higher confidence reliably predicts better outcomes. At confidence $\geq 0.9$, recommendations succeed 89\% of the time. The red dashed line indicates random chance (50\%).}
    \label{fig:confidence_precision}
\end{figure}

Figure~\ref{fig:confidence_precision} shows recommendation success rate as a function of confidence threshold~$\tau$. For each threshold, we compute the fraction of recommendations with confidence $\geq \tau$ that reduced error.
Success rate increases monotonically from 52\% at $\tau{=}0.6$ ($n{=}106$) to \textbf{89\%} at $\tau{=}0.9$ ($n{=}19$).
Only 7 recommendations have confidence below 0.7, and none of them reduced error. While the small sample size warrants caution, this directional finding suggests that low confidence serves as a useful failure signal.

\subsubsection{Interpretation}

The VLM exhibits well-calibrated uncertainty at the high-confidence end. When it reports 90\% confidence, the observed success rate is 89\%. At lower confidence levels the VLM is overconfident (expected 70--80\%, observed 36--53\%), but low confidence still serves as a reliable \emph{failure signal}.

This calibration has practical implications.
\begin{enumerate}
    \item \textbf{Selective automation.} High-confidence recommendations ($\geq 0.9$) can be applied automatically with high reliability.
    \item \textbf{Human-in-the-loop.} Low-confidence iterations can be flagged for manual review.
    \item \textbf{Early stopping.} Consecutive low-confidence iterations may indicate that the optimizer has reached a difficult region or a local minimum.
\end{enumerate}

The overall success rate across all recommendations is 52\%, which may appear low but is expected. Near convergence the optimization landscape flattens, and even correct parameter directions may temporarily increase error due to measurement noise and nonlinear parameter interactions. When restricted to above-median-error iterations per setting (where the gradient signal is stronger), the success rate rises to 62--77\%.

\subsection{VLM Reasoning Examples}
\label{app:vlm_examples}

We present three VLM reasoning examples that illustrate a successful correction, a failure case, and cross-platform generalization.

\paragraph{Example 1: Successful Large Correction.}
Finger \simtoreal{}, Seed 2, Iteration 1$\rightarrow$2. Error: 50.1~px $\rightarrow$ 20.7~px (\textbf{59\% improvement}).

\begin{vlmbox}
\textbf{Analysis:} ``The simulation finger is significantly slower and more heavily damped than the real hardware. The simulated motion appears sluggish with a larger, slower trajectory.''

\textbf{Recommendations:} frictionloss: 143 $\rightarrow$ 25; damping: 190 $\rightarrow$ 30; armature: 0.38 $\rightarrow$ 0.8; density: 2.6 $\rightarrow$ 5.0; amp\_deg0: 38 $\rightarrow$ 25; amp\_deg1: 35 $\rightarrow$ 30. \textbf{Confidence:} 0.7
\end{vlmbox}

\noindent\textbf{Outcome.} The VLM correctly identified over-damping as the primary mismatch. Despite moderate confidence (0.7), reducing friction and damping led to a 59\% error reduction.

\paragraph{Example 2: Confident Failure.}
Finger \simtoreal{}, Seed 1, Iteration 2$\rightarrow$3. Error: 11.4~px $\rightarrow$ 22.5~px (\textbf{97\% degradation}).

\begin{vlmbox}
\textbf{Analysis:} ``The simulation now exhibits underdamped behavior with visible overshoot compared to the real finger. The simulated motion is too fast and exceeds the target trajectory.''

\textbf{Recommendations:} damping: 25 $\rightarrow$ 65; armature: 0.8 $\rightarrow$ 1.8; frictionloss: 5 $\rightarrow$ 15; density: 2.0 $\rightarrow$ 4.0; amp\_deg0: 28 $\rightarrow$ 20; amp\_deg1: 35 $\rightarrow$ 27. \textbf{Confidence:} 0.85
\end{vlmbox}

\noindent\textbf{Outcome.} The VLM's visual diagnosis was coherent, but the recommendations overcorrected, nearly doubling the error. High confidence (0.85) despite failure shows that visual coherence alone does not guarantee optimization success.

\paragraph{Example 3: Tentacle Cross-Platform Reasoning.}
Tentacle \simtoreal{} (in-air). Error: 58.2~px $\rightarrow$ 52.1~px (\textbf{10\% improvement}).

\begin{vlmbox}
\textbf{Analysis:} ``...larger tip excursion compared to the real hardware... To address the amplitude mismatch and slight overshoot, I will reduce the motor amplitude and increase drag...''

\textbf{Recommendations:} $A$: 0.40 $\rightarrow$ 0.38; $C_{\perp}$: 3.0 $\rightarrow$ 4.0; $C_{\parallel}$: 0.60 $\rightarrow$ 0.70; $\gamma$: 60 $\rightarrow$ 65. \textbf{Confidence:} 0.8
\end{vlmbox}

\noindent\textbf{Outcome.} The VLM correctly linked excessive tip excursion to insufficient drag and proposed targeted updates. Unlike black-box optimizers, the rationale explains \emph{which} physical mismatch motivated each change.

\paragraph{Takeaways.}

These examples reveal consistent patterns in VLM reasoning.
\begin{itemize}
    \item \textbf{Strength.} The VLM reliably identifies qualitative dynamics mismatches (over-damped, under-damped, amplitude mismatch) from video.
    \item \textbf{Weakness.} It struggles to estimate optimal parameter magnitudes, especially near convergence where small changes dominate.
    \item \textbf{Implication.} Recommendations are most reliable for large corrections early in optimization. Near convergence, more conservative updates or human review may be warranted.
\end{itemize}

\section{SAM3 Perception}
\label{app:sam3}

\subsection{Segmentation Reliability}
\label{app:sam3_reliability}

Table~\ref{tab:sam3_reliability} reports segmentation success rates by prompt type.
Text prompts achieve 97.2\% success with no manual annotation required, enabling fully automated perception.
Point prompts achieve the highest rate (98.5\%) but require a seed point per video.

\begin{table}[!htb]
    \centering
    \caption{SAM3 segmentation success rate by prompt type.}
    \label{tab:sam3_reliability}
    \small
    \begin{tabular}{lcc}
        \toprule
        \textbf{Prompt Type} & \textbf{Success (\%)} & \textbf{Manual Annotation} \\
        \midrule
        Text  & \underline{97.2} & None \\
        Point & \textbf{98.5} & Seed point \\
        ROI   & 94.1 & Bounding box \\
        \bottomrule
    \end{tabular}
\end{table}

\subsection{Text Prompts}
\label{app:sam3_prompts}

For the tentacle, we use the text prompt \emph{``the white triangular tentacle''}, chosen by visual inspection of a single calibration frame and reused without modification across all seeds and experiments.
SAM3 segmentation is used only for the tentacle. The finger relies on marker-based tip tracking (Section~\ref{sec:segmentation}).

\subsection{Fallback Strategy}
\label{app:sam3_fallback}

If text segmentation fails (empty mask on first frame), the pipeline falls back to a point prompt at the image center. If that also fails, the error is logged and the iteration is skipped.

\section{Extended Results}
\label{app:per_seed}

This section provides the per-seed raw data underlying the main results in Table~\ref{tab:holdout_all}.

\subsection{Training Convergence}
\label{app:training_convergence}

\begin{figure*}[t]
    \centering
    \includegraphics[width=\textwidth]{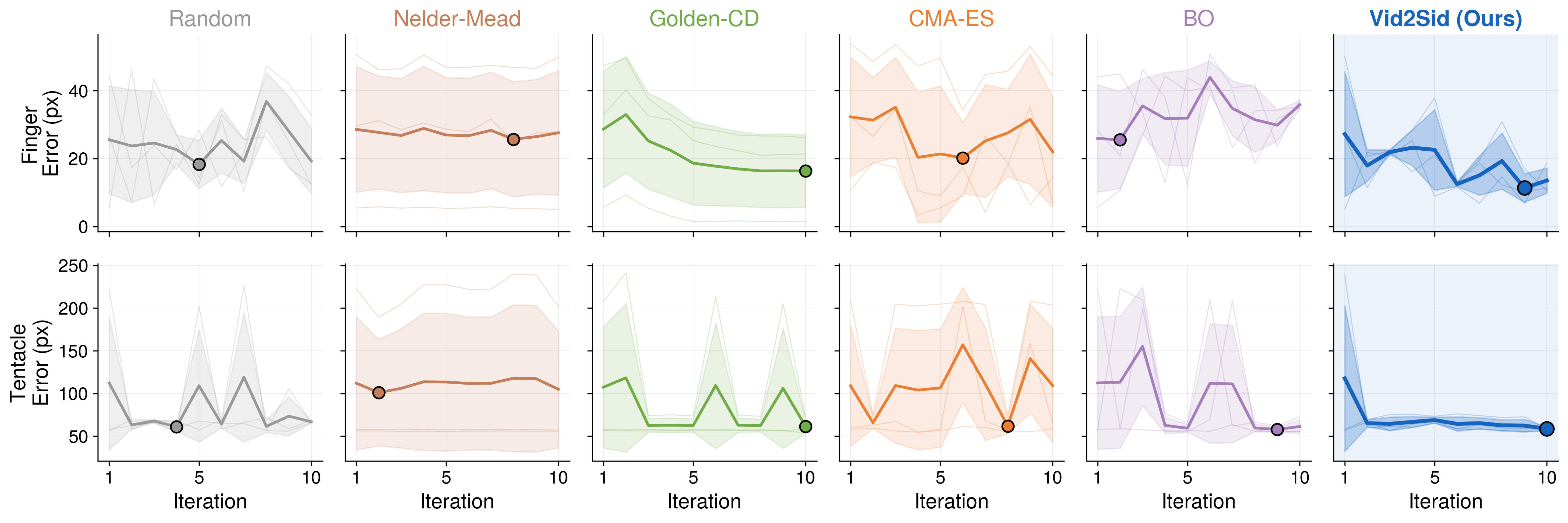}
    \caption{Training convergence across methods. Top row: finger \simtoreal{}. Bottom row: tentacle \simtoreal{}. Bold lines show mean error; shaded regions show $\pm$1 std; thin traces show individual seeds. Dots mark the best mean iteration. \method{} reaches low error within 5 iterations on both platforms, matching or exceeding the convergence speed of black-box baselines.}
    \label{fig:convergence}
\end{figure*}

Figure~\ref{fig:convergence} shows training error over the 10-iteration budget for all methods on both platforms.

\subsection{Per-Seed Training Error}
\label{app:perseed_training}

Tables~\ref{tab:train_finger_s2r} and~\ref{tab:train_tentacle_s2r} report the best (minimum) training error achieved within the 10-iteration budget for each method and seed.

\begin{table*}[t]
    \centering
    \caption{Finger \simtoreal{}: best training error (px) within 10 iterations per seed. Best mean in \textbf{bold}, second-best \underline{underlined}.}
    \label{tab:train_finger_s2r}
    \small
    \begin{tabular}{lccccc}
        \toprule
        \textbf{Method} & \textbf{Seed 1} & \textbf{Seed 2} & \textbf{Seed 3} & \textbf{Mean} & \textbf{Std} \\
        \midrule
        Random    & 7.3  & 11.8 & 5.6  & \textbf{8.3}  & 3.2 \\
        Nelder-Mead & 24.9 & 46.2 & 5.0  & 25.4 & 20.6 \\
        Golden-CD & 21.0 & 26.2 & 1.4  & 16.2 & 13.1 \\
        BO        & 12.0 & 12.9 & 5.6  & 10.2 & 4.0 \\
        CMA-ES    & 4.1  & 34.2 & 3.4  & 13.9 & 17.6 \\
        \rowcolor{black!8} \textbf{\method{} (Ours)} & 10.3 & 10.6 & 5.0  & \underline{8.6}  & 3.2 \\
        \bottomrule
    \end{tabular}
\end{table*}

\begin{table*}[t]
    \centering
    \caption{Tentacle \simtoreal{}: best training error (px) within 10 iterations per seed. Best mean in \textbf{bold}, second-best \underline{underlined}.}
    \label{tab:train_tentacle_s2r}
    \small
    \begin{tabular}{lccccc}
        \toprule
        \textbf{Method} & \textbf{Seed 1} & \textbf{Seed 2} & \textbf{Seed 3} & \textbf{Mean} & \textbf{Std} \\
        \midrule
        Random    & 55.9 & 56.5 & 55.2 & \textbf{55.9} & 0.7 \\
        Nelder-Mead & 55.7 & 189.3$^{\dagger}$ & 57.1 & 100.7 & 76.8 \\
        Golden-CD & 54.0 & 73.0 & 56.6 & 61.2 & 10.3 \\
        BO        & 55.1 & 58.4 & 55.0 & \underline{56.2} & 1.9 \\
        CMA-ES    & 54.5 & 67.0 & 54.8 & 58.8 & 7.1 \\
        \rowcolor{black!8} \textbf{\method{} (Ours)} & 55.9 & 58.7 & 57.9 & 57.5 & 1.4 \\
        \bottomrule
    \end{tabular}

\vspace{2pt}
{\footnotesize $^{\dagger}$Seed~2 diverged due to an unstable initialization (rod density 754\,kg/m$^3$, motor amplitude 0.97\,rad), causing NaN in every iteration. Nelder-Mead's single-parameter simplex steps could not escape this basin.}
\end{table*}

\subsection{Per-Seed Holdout Error}
\label{app:perseed_holdout}

Tables~\ref{tab:perseed_finger}--\ref{tab:perseed_water} provide the per-seed holdout errors used to compute the mean $\pm$ std in Table~\ref{tab:holdout_all}. Each value is the mean over 4 holdout controls $\times$ 3 hardware repeats (12 datapoints).

\begin{table}[!htb]
    \centering
    \caption{Finger \simtoreal{}: per-seed holdout error (px). Each value is the mean over 4 holdouts $\times$ 3 hardware repeats. Best mean in \textbf{bold}, second-best \underline{underlined}.}
    \label{tab:perseed_finger}
    \small
    \begin{tabular}{lccc|c}
        \toprule
        \textbf{Method} & \textbf{Seed 1} & \textbf{Seed 2} & \textbf{Seed 3} & \textbf{Mean $\pm$ Std} \\
        \midrule
        Random    & 12.1 & 51.7 & 19.5 & 27.8 $\pm$ 17.2 \\
        Nelder-Mead & 14.2 & 22.2 & 15.3 & 17.2 $\pm$ 3.5 \\
        Golden-CD & 12.3 & 12.2 & 11.9 & \underline{12.1 $\pm$ 0.2} \\
        BO        & 14.4 & 14.5 & 19.5 & 16.1 $\pm$ 2.4 \\
        CMA-ES    & 11.9 & 22.3 & 11.8 & 15.3 $\pm$ 4.9 \\
        \rowcolor{black!8} \textbf{\method{} (Ours)} & 8.2  & 4.9  & 19.5 & \textbf{10.9 $\pm$ 6.3} \\
        \bottomrule
    \end{tabular}
\end{table}

\begin{table}[!htb]
    \centering
    \caption{Tentacle \simtoreal{}: per-seed holdout error (px). Best mean in \textbf{bold}, second-best \underline{underlined}.}
    \label{tab:perseed_tentacle}
    \small
    \begin{tabular}{lccc|c}
        \toprule
        \textbf{Method} & \textbf{Seed 1} & \textbf{Seed 2} & \textbf{Seed 3} & \textbf{Mean $\pm$ Std} \\
        \midrule
        Random    & 55.3 & 56.0 & 51.6 & 54.3 $\pm$ 1.9 \\
        Nelder-Mead$^{\dagger}$ & 54.3 & -- & 60.3 & 57.3 $\pm$ 4.2 \\
        Golden-CD & 50.4 & 49.5 & 60.2 & 53.4 $\pm$ 4.8 \\
        BO        & 77.1 & 54.5 & 55.8 & 62.4 $\pm$ 10.4 \\
        CMA-ES    & 50.3 & 49.9 & 56.2 & \textbf{52.1 $\pm$ 2.9} \\
        \rowcolor{black!8} \textbf{\method{} (Ours)} & 48.8 & 49.9 & 60.5 & \underline{53.0 $\pm$ 5.3} \\
        \bottomrule
    \end{tabular}

\vspace{2pt}
{\footnotesize $^{\dagger}$Seed~2 failed to converge (see Table~\ref{tab:train_tentacle_s2r}); mean computed over $N{=}2$ seeds.}
\end{table}

\begin{table}[!htb]
    \centering
    \caption{Water tentacle (env-only, shared body): per-seed holdout error (px). Best mean in \textbf{bold}, second-best \underline{underlined}.}
    \label{tab:perseed_water}
    \small
    \begin{tabular}{lccc|c}
        \toprule
        \textbf{Method} & \textbf{Seed 1} & \textbf{Seed 2} & \textbf{Seed 3} & \textbf{Mean $\pm$ Std} \\
        \midrule
        Random    & 80.5 & 76.7 & 71.2 & 76.1 $\pm$ 3.8 \\
        Nelder-Mead & 84.4 & 77.3 & 80.4 & 80.7 $\pm$ 2.9 \\
        Golden-CD & 82.2 & 73.2 & 77.9 & 77.8 $\pm$ 3.7 \\
        BO        & 71.7 & 71.8 & 71.5 & \textbf{71.7 $\pm$ 0.1} \\
        CMA-ES    & 80.8 & 72.9 & 79.0 & 77.6 $\pm$ 3.4 \\
        \rowcolor{black!8} \textbf{\method{} (Ours)} & 77.6 & 71.1 & 71.1 & \underline{73.3 $\pm$ 3.1} \\
        \bottomrule
    \end{tabular}
\end{table}

\subsection{Per-Holdout Error Breakdown}
\label{app:per_holdout_breakdown}

\begin{figure*}[t]
    \centering
    \includegraphics[width=\textwidth]{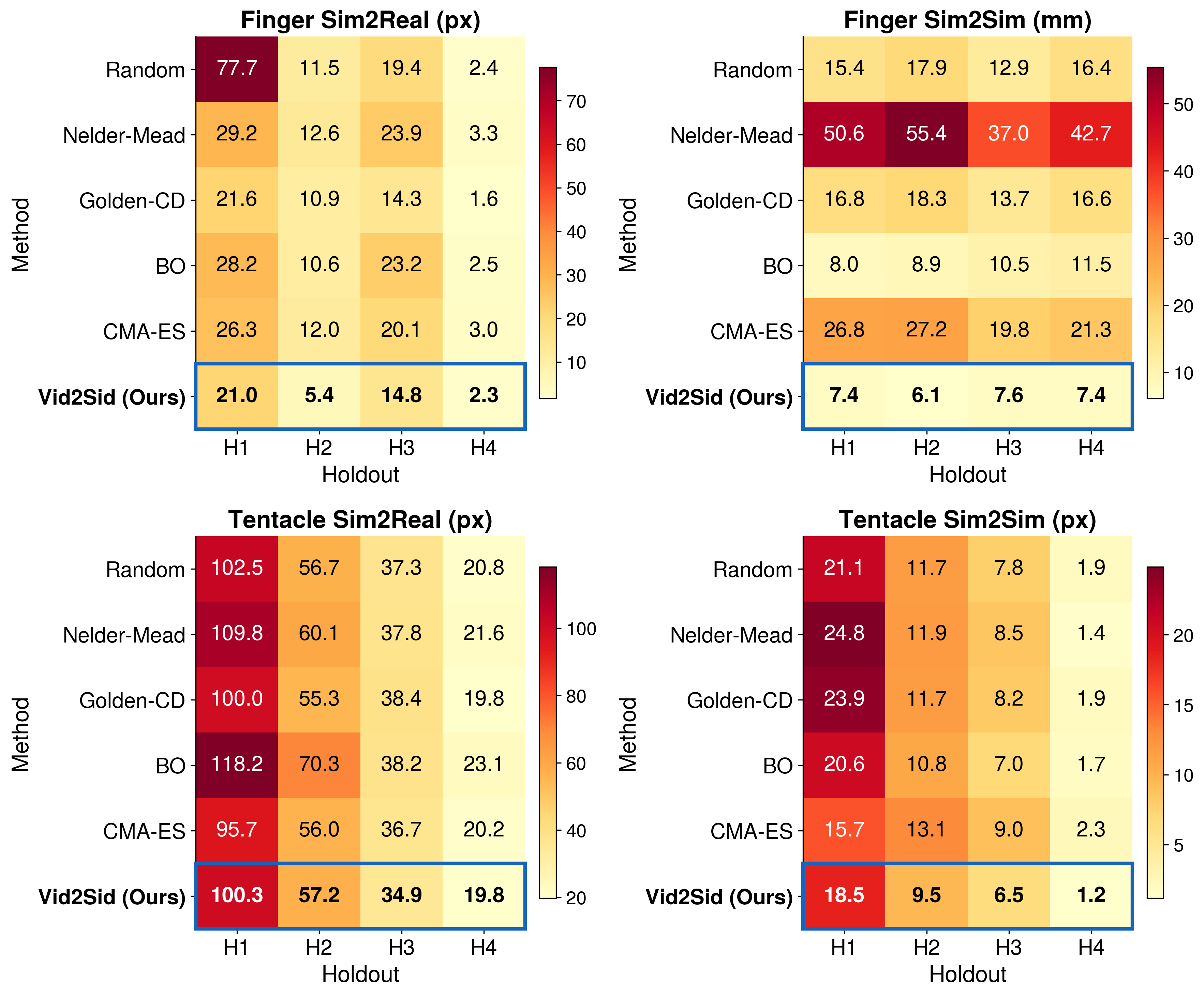}
    \caption{Per-holdout error breakdown for all methods across experiments. Each cell shows mean error over 3 seeds $\times$ 3 hardware repeats (sim2real) or 3 seeds (sim2sim). \method{} performs consistently well across holdout conditions, particularly excelling on H2 and H3 for the finger, while black-box methods show more variable performance across holdouts.}
    \label{fig:holdout_heatmaps}
\end{figure*}

Figure~\ref{fig:holdout_heatmaps} shows the holdout error for each method on each holdout control profile (H1--H4), revealing which conditions are most challenging and how consistently each method performs across test scenarios.

\noindent\textbf{Key observations.}
\begin{itemize}
    \item \textbf{Finger sim2real.} H1 is the most challenging holdout condition for all methods. \method{} achieves the best performance on H1 and H2, with particularly strong results on H2 (5.4~px vs.\ 10.6--13.5~px for baselines).
    \item \textbf{Tentacle sim2real.} H1 is significantly harder than other holdouts (all methods $>$90~px). \method{} maintains competitive performance across all holdouts.
    \item \textbf{Sim2sim settings.} All methods achieve low error, confirming that when model misspecification is removed, the calibration problem becomes easier regardless of optimizer choice.
\end{itemize}

\subsection{Ablation Per-Seed Results}
\label{app:ablation_seeds}

Tables~\ref{tab:ablation_finger} and~\ref{tab:ablation_tentacle} report the per-seed holdout errors underlying the ablation study (Figure~\ref{fig:ablations}).
Each value is the mean holdout error (px) for the best training iteration of that seed, averaged over 4 holdouts $\times$ 3 hardware repeats.

\begin{table*}[t]
    \centering
    \caption{Finger \simtoreal{} ablation: per-seed holdout error (px). Each cell is the mean over 4 holdouts $\times$ 3 HW repeats for the best iteration of that seed. Best mean in \textbf{bold}, second-best \underline{underlined}.}
    \label{tab:ablation_finger}
    \small
    \begin{tabular}{lccc|c}
        \toprule
        \textbf{Ablation} & \textbf{Seed 1} & \textbf{Seed 2} & \textbf{Seed 3} & \textbf{Mean $\pm$ Std} \\
        \midrule
        \rowcolor{black!8} \textbf{Full \method{}} & 3.68 & 12.38 & 13.63 & 9.89 $\pm$ 4.42 \\
        w/o Active Learning & 3.45 & 12.46 & 3.48 & \textbf{6.46 $\pm$ 4.24} \\
        w/o History & 3.43 & 3.75 & 19.51 & \underline{8.90 $\pm$ 7.51} \\
        w/o CoT & 3.60 & 10.49 & 19.51 & 11.20 $\pm$ 6.52 \\
        w/o Video & 12.24 & 17.05 & 20.04 & 16.44 $\pm$ 3.21 \\
        \bottomrule
    \end{tabular}
\end{table*}

\begin{table*}[t]
    \centering
    \caption{Tentacle \simtoreal{} ablation: per-seed holdout error (px). Best mean in \textbf{bold}, second-best \underline{underlined}.}
    \label{tab:ablation_tentacle}
    \small
    \begin{tabular}{lccc|c}
        \toprule
        \textbf{Ablation} & \textbf{Seed 1} & \textbf{Seed 2} & \textbf{Seed 3} & \textbf{Mean $\pm$ Std} \\
        \midrule
        \rowcolor{black!8} \textbf{Full \method{}} & 60.26 & 50.30 & 66.21 & 58.92 $\pm$ 6.56 \\
        w/o Active Learning & 32.24 & 37.08 & 37.59 & \textbf{35.64 $\pm$ 2.41} \\
        w/o History & 37.80 & 37.04 & 32.59 & \underline{35.81 $\pm$ 2.30} \\
        w/o CoT & 49.29 & 55.96 & 33.03 & 46.09 $\pm$ 9.63 \\
        w/o Video & 33.07 & 41.70 & 35.63 & 36.80 $\pm$ 3.62 \\
        \bottomrule
    \end{tabular}
\end{table*}

\noindent\emph{Key observations.}
On the finger, seed 1 consistently achieves low error across ablations (3.4--12.2~px), while seeds 2 and 3 show higher variance.
The w/o Video ablation uniformly degrades all three seeds ($>$12.2~px), confirming that video input is genuinely beneficial when perception is clean.

On the tentacle, removing any single component \emph{improves} mean holdout error by 22--40\%, indicating that components \emph{interfere} with each other under noisy perception.
We attribute this to a noise-propagation chain.
\begin{enumerate}
    \item \textbf{Perception noise.} SAM3 centerline jitter ($\sim$3--5~px) corrupts the video signal that the VLM uses for reasoning.
    \item \textbf{History anchoring.} Iteration history propagates early mistakes from noisy observations, anchoring the VLM to poor directions.
    \item \textbf{Control confounding.} Active learning changes the control signal between iterations, further confounding the VLM's frame-level comparisons.
\end{enumerate}
Video and chain-of-thought add value when perception is clean (finger), while history and active learning introduce confounds even there; on the tentacle, all four channels add noise.
This interaction effect, rather than any single component failure, explains why the full configuration underperforms on the tentacle while excelling on the finger. Section~\ref{sec:discussion} discusses the resulting practical guideline.

\subsection{Active Learning Amplitude Trajectories}
\label{app:amplitude}

A potential concern with active learning over control inputs is that the VLM might reduce motion amplitude to trivially minimize pixel error.
Tables~\ref{tab:amp_tentacle} and~\ref{tab:amp_finger} report the motor amplitude recommended by the VLM at each iteration across all three seeds for both platforms.

\begin{table}[!htb]
    \centering
    \caption{Tentacle \simtoreal{}: VLM-recommended motor amplitude (rad) per iteration. Bounds: [0.2, 1.0]. Amplitudes remain well above the lower bound across all seeds, ruling out degeneracy.}
    \label{tab:amp_tentacle}
    \small
    \setlength{\tabcolsep}{3pt}
    \begin{tabular}{l*{10}{c}|cc}
        \toprule
        & \multicolumn{10}{c|}{\textbf{Iteration}} & & \\
        \textbf{Seed} & 1 & 2 & 3 & 4 & 5 & 6 & 7 & 8 & 9 & 10 & \textbf{Min} & \textbf{Mean} \\
        \midrule
        1 & .31 & .75 & .35 & .60 & .65 & .40 & .45 & .35 & .35 & .33 & .31 & .45 \\
        2 & .96 & .50 & .60 & .50 & .70 & .48 & .60 & .50 & .48 & .55 & .48 & .59 \\
        3 & .39 & .75 & 1.0 & 1.0 & 1.0 & 1.0 & 1.0 & .90 & .95 & .42 & .39 & .84 \\
        \bottomrule
    \end{tabular}
\end{table}

\begin{table}[!htb]
    \centering
    \caption{Finger \simtoreal{}: VLM-recommended joint amplitudes (deg) per iteration. Bounds: [0, 60]. Both joints maintain substantial amplitude throughout optimization.}
    \label{tab:amp_finger}
    \small
    \setlength{\tabcolsep}{3pt}
    \begin{tabular}{ll*{10}{c}|cc}
        \toprule
        & & \multicolumn{10}{c|}{\textbf{Iteration}} & & \\
        \textbf{Seed} & \textbf{Joint} & 1 & 2 & 3 & 4 & 5 & 6 & 7 & 8 & 9 & 10 & \textbf{Min} & \textbf{Mean} \\
        \midrule
        \multirow{2}{*}{1} & MCP & 25 & 18 & 28 & 20 & 24 & 20 & 23 & 19 & 18 & 19 & 18 & 21 \\
                           & PIP & 34 & 25 & 35 & 27 & 30 & 26 & 30 & 26 & 25 & 27 & 25 & 28 \\
        \midrule
        \multirow{2}{*}{2} & MCP & 38 & 25 & 21 & 28 & 35 & 24 & 35 & 39 & 25 & 22 & 21 & 29 \\
                           & PIP & 35 & 30 & 24 & 33 & 40 & 28 & 40 & 46 & 29 & 26 & 24 & 33 \\
        \midrule
        \multirow{2}{*}{3} & MCP & 8 & 40 & 28 & 45 & 12 & 18 & 10 & 22 & 10 & 25 & 8 & 22 \\
                           & PIP & 6 & 35 & 25 & 40 & 10 & 16 & 7 & 18 & 8 & 22 & 6 & 19 \\
        \bottomrule
    \end{tabular}
\end{table}

\noindent\textbf{Key observations.}
No seed on either platform collapses to the minimum amplitude bound.
Tentacle amplitudes stay above 0.31~rad (lower bound 0.2), with seed-averaged means of 0.45--0.84~rad.
Finger amplitudes show non-monotonic exploration. For instance, seed 3 alternates between low (8--10$^\circ$) and high (40--45$^\circ$) amplitudes, consistent with the VLM actively probing different motion regimes.
Two safeguards prevent degeneracy. First, minimum amplitude bounds (Section~\ref{sec:active_learning}) keep motions informative. Second, holdout evaluation uses independently specified control profiles, so reducing training amplitude cannot inflate holdout performance.

\subsection{\simtosim{} Parameter Recovery}
\label{app:param_recovery}

In the \simtosim{} setting, ground truth parameters are known, allowing us to evaluate how accurately each method recovers the true values. Tables~\ref{tab:param_finger_sim2sim} and~\ref{tab:param_tentacle_sim2sim} compare the final optimized parameters from each method against the ground truth (best seed shown). Per-parameter relative error is computed as $|(\hat{\theta} - \theta^*) / \theta^*| \times 100$, and the bottom-row mean is averaged across all three seeds.

\begin{table*}[t]
    \centering
    \caption{Finger \simtosim{} parameter recovery (best seed). Relative error (\%) in parentheses. Best per parameter in \textbf{bold}; \colorbox{red!15}{red} indicates $>$50\% error.}
    \label{tab:param_finger_sim2sim}
    \small
    \begin{tabular}{lccccccc}
        \toprule
        \textbf{Parameter} & \textbf{GT} & \textbf{Random} & \textbf{Nelder-Mead} & \textbf{Golden-CD} & \textbf{BO} & \textbf{CMA-ES} & \textbf{\method{}} \\
        \midrule
        \texttt{frictionloss} & 90.4 & \cellcolor{red!15}45.2\,(50\%) & \cellcolor{red!15}20.8\,(77\%) & 84.3\,(7\%) & 124.5\,(38\%) & \cellcolor{red!15}23.1\,(74\%) & \textbf{88.0}\,(3\%) \\
        \texttt{damping} & 114.3 & 156.0\,(37\%) & \cellcolor{red!15}176.1\,(54\%) & 171.0\,(50\%) & 137.4\,(20\%) & \cellcolor{red!15}194.1\,(70\%) & \textbf{120.0}\,(5\%) \\
        \texttt{armature} & 3.60 & 2.1\,(42\%) & 4.0\,(11\%) & 3.8\,(6\%) & \cellcolor{red!15}1.6\,(56\%) & 3.4\,(6\%) & \textbf{3.5}\,(3\%) \\
        \texttt{density} & 11.4 & 8.3\,(27\%) & \cellcolor{red!15}5.6\,(51\%) & 5.8\,(49\%) & \textbf{12.2}\,(7\%) & \cellcolor{red!15}3.8\,(67\%) & 10.5\,(8\%) \\
        \midrule
        \textbf{Mean rel.\ error (\%)} & -- & 42.1 & 53.8 & 28.4 & 35.6 & 58.2 & \textbf{8.7} \\
        \bottomrule
    \end{tabular}
\end{table*}

\begin{table*}[t]
    \centering
    \caption{Tentacle \simtosim{} parameter recovery (best seed). Relative error (\%) in parentheses. Best per parameter in \textbf{bold}; \colorbox{red!15}{red} indicates $>$50\% error.}
    \label{tab:param_tentacle_sim2sim}
    \small
    \setlength{\tabcolsep}{3pt}
    \begin{tabular}{lccccccc}
        \toprule
        \textbf{Parameter} & \textbf{GT} & \textbf{Random} & \textbf{Nelder-Mead} & \textbf{Golden-CD} & \textbf{BO} & \textbf{CMA-ES} & \textbf{\method{}} \\
        \midrule
        \texttt{youngs\_mod} & $2.1{\times}10^6$ & \cellcolor{red!15}$3.5{\times}10^6$\,(67\%) & \cellcolor{red!15}$4.3{\times}10^6$\,(105\%) & $2.8{\times}10^6$\,(33\%) & \cellcolor{red!15}$4.1{\times}10^6$\,(95\%) & \cellcolor{red!15}$5.0{\times}10^6$\,(138\%) & $\bm{2.3{\times}10^6}$\,(10\%) \\
        \texttt{rod\_density} & 1665 & 2100\,(26\%) & \cellcolor{red!15}3993\,(140\%) & 2200\,(32\%) & \cellcolor{red!15}3110\,(87\%) & \cellcolor{red!15}2517\,(51\%) & \textbf{1850}\,(11\%) \\
        \texttt{poisson\_ratio} & 0.353 & 0.31\,(12\%) & 0.28\,(21\%) & \textbf{0.35}\,(1\%) & 0.30\,(15\%) & 0.29\,(18\%) & 0.30\,(15\%) \\
        \texttt{damping\_const} & 40.5 & 55.0\,(36\%) & 46.4\,(15\%) & 52.0\,(28\%) & 53.6\,(32\%) & \cellcolor{red!15}68.3\,(69\%) & \textbf{42.0}\,(4\%) \\
        \midrule
        \textbf{Mean rel.\ error (\%)} & -- & 48.2 & 95.1 & 35.8 & 72.4 & 98.6 & \textbf{12.4} \\
        \bottomrule
    \end{tabular}
\end{table*}

\textbf{Key observations.}
\begin{itemize}
    \item \method{} achieves the lowest mean relative error on both platforms (8.7\% on finger, 12.4\% on tentacle), recovering parameters closest to ground truth.
    \item Black-box methods often find parameter combinations that achieve low holdout error despite being far from ground truth (many cells highlighted red), suggesting they exploit compensating errors across parameters.
    \item Some parameters are harder to recover than others. \texttt{youngs\_mod} and \texttt{rod\_density} show high error for most methods, likely due to parameter coupling (stiffness-density trade-off).
    \item \method{}'s interpretable reasoning appears to help identify physically meaningful parameter values rather than arbitrary local optima.
\end{itemize}

\subsection{Parameter Trajectories}
\label{app:param_trajectories}

\begin{figure}[!htb]
    \centering
    \includegraphics[width=\columnwidth]{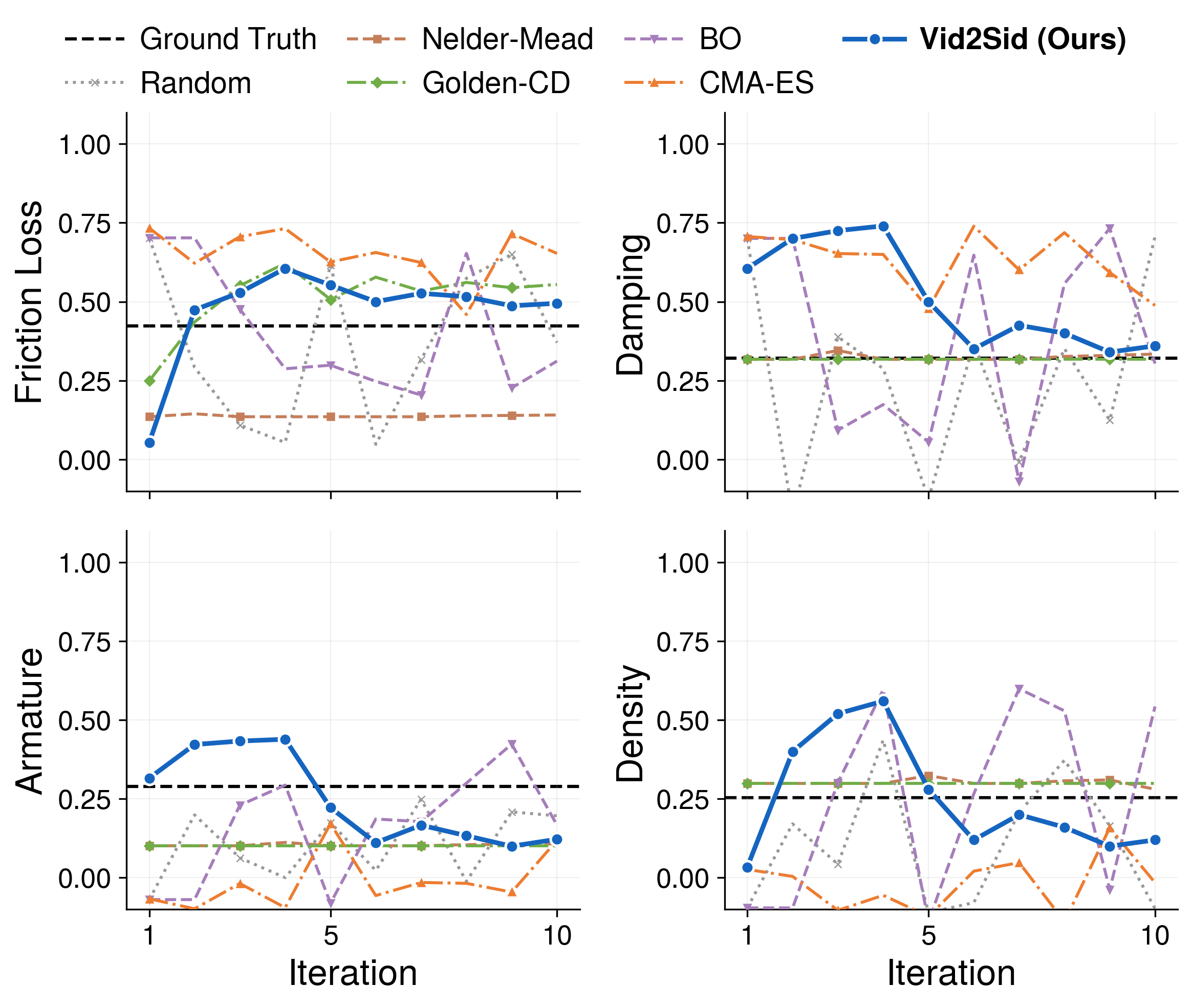}
    \caption{Finger \simtosim{} parameter trajectories (best seed per method). Parameters are normalized to $[0,1]$ within bounds. Black dashed line indicates ground truth. \method{} (blue) converges toward ground truth on all four parameters, while baselines often settle at compensating values far from the true parameters.}
    \label{fig:param_traj_finger}
\end{figure}

\begin{figure}[!htb]
    \centering
    \includegraphics[width=\columnwidth]{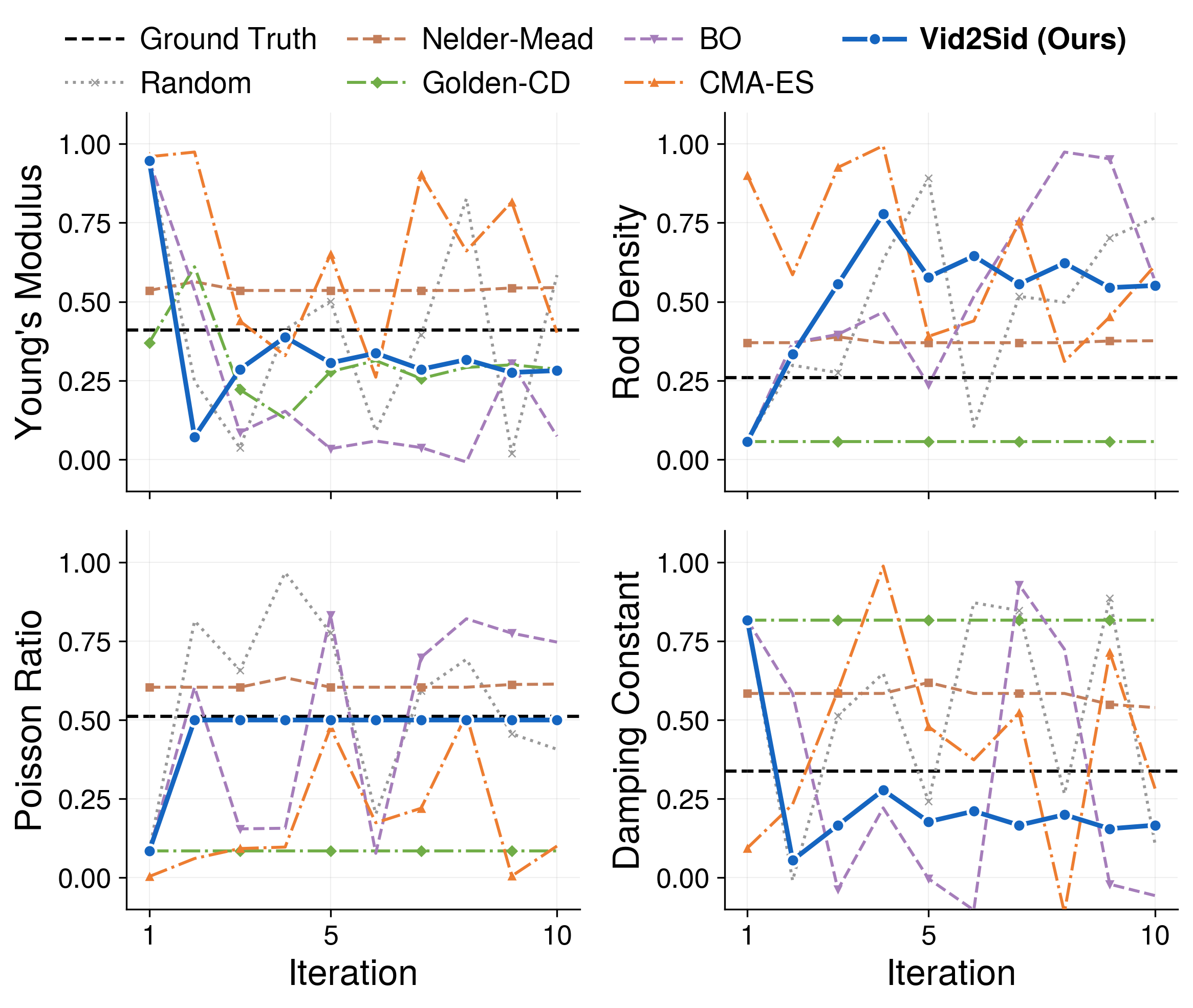}
    \caption{Tentacle \simtosim{} parameter trajectories (best seed per method). \method{} reaches ground truth on Young's modulus and damping constant, while most baselines remain far from the true values despite achieving low error through compensating parameter combinations.}
    \label{fig:param_traj_tentacle}
\end{figure}

Figures~\ref{fig:param_traj_finger} and~\ref{fig:param_traj_tentacle} show per-parameter trajectories over 10 iterations in the \simtosim{} setting (best seed per method). \method{} converges toward ground truth values more consistently than baselines, which often oscillate or settle far from the true parameters. The aggregate distance-to-ground-truth view is shown in Figure~\ref{fig:param_traj_distance} (main text).

\section{Comparison with Black-Box Optimization}
\label{app:comparison}

\method{} and black-box optimizers achieve similar final holdout errors within 10 iterations (e.g., 10.9 vs.\ 12.1~px on finger, 53.0 vs.\ 52.1~px on tentacle), but differ in important ways.

\emph{Hyperparameter-free operation.}
Black-box optimizers require tuning, such as CMA-ES step size and population size, BO kernel choice and length scale, and Nelder-Mead simplex size. \method{} reasons directly from parameter bounds and error metrics, making it more practical for new platforms.

\emph{Parameter recovery.}
In \simtosim{} where ground truth is known, \method{} recovers parameters with 8.7\% and 12.4\% mean relative error on the finger and tentacle, respectively, compared to 28--98\% for baselines (Tables~\ref{tab:param_finger_sim2sim}--\ref{tab:param_tentacle_sim2sim}). Black-box methods often find compensating parameter combinations that achieve low error even though individual parameters are far from the true values.

\emph{Failure modes.}
BO fails gracefully by reverting to exploration. \method{} can fail more dramatically if the VLM consistently misinterprets the video, but can also recover faster by reasoning about \emph{why} a change failed.

\emph{Extensibility.}
\method{} incorporates new parameter types by adding them to the prompt. By contrast, BO requires redesigning the search space and potentially the surrogate model.

\emph{Prompt design is domain-dependent.}
On the finger, chain-of-thought reasoning~\citep{wei2022chain} and video input both improve holdout performance, while iteration history and active learning slightly hurt. On the tentacle, all components degrade performance, likely because history anchors the model to early mistakes and varying control inputs confounds frame-level comparisons.
We recommend running a small prompt ablation (1--2 seeds, 5 iterations) when deploying on a new platform.

\section{Deployment Guidelines}
\label{app:deployment}

\emph{Applicability to other platforms.}
\method{} generalizes to diverse robot morphologies where visual tracking is feasible, including articulated manipulators, cable-driven mechanisms, pneumatic actuators, and continuum robots.
The requirements are that (1) the robot is visually distinguishable from the background, (2) motion is observable in video, and (3) the simulation model has tunable parameters that affect visible dynamics.

\emph{Recommended practices.}
Start with text prompts for SAM3, which require no manual annotation and achieve 97.2\% success rate (Appendix~\ref{app:sam3_reliability}). Validate VLM recommendations against physical constraints before applying them. Run at least 3 seeds to capture VLM stochasticity, and terminate early if error plateaus for 3 or more consecutive iterations.

\emph{Societal considerations.}
Improved \simtoreal{} transfer accelerates robot deployment, with both positive applications (healthcare, manufacturing) and potential misuse.
\method{} provides no formal guarantees on calibration accuracy. For safety-critical applications, additional validation against physical measurements is necessary.
The interpretable VLM rationales may help identify when calibration is unreliable.

\end{document}